%% file: main.tex
\documentclass[sigconf,nonacm]{acmart}
\AtBeginDocument{%
  }

\renewcommand\footnotetextcopyrightpermission[1]{}

\usepackage{graphicx}
\usepackage{amsmath}
\usepackage{booktabs}
\usepackage{multirow}
\usepackage{enumitem}
\usepackage{tabularx}
\usepackage{xcolor}
\usepackage[normalem]{ulem}
\usepackage[dvipsnames]{xcolor}
\usepackage{capt-of}
\usepackage{adjustbox}
\usepackage{lineno}
\usepackage{pifont}  % for xmark
\usepackage{makecell} % for line breaks in headers
\usepackage[table]{xcolor}

\newcommand{\xmark}{\ding{55}}

\newcommand{\blue}[1]{\textcolor{blue}{#1}}
\renewcommand{\blue}[1]{#1}

\definecolor{chartmodels}{RGB}{255,255,255}
\definecolor{opensource}{RGB}{255,255,255}
\definecolor{finetuned}{RGB}{255,255,255}
\definecolor{closedsource}{RGB}{255,255,255}
\definecolor{baseline}{RGB}{255,255,255}

\newcommand{\good}[1]{\textcolor{green!60!black}{#1}}
\newcommand{\bad}[1]{\textcolor{red!70!black}{#1}}

\setcopyright{none}

\begin{document}

\title{ThermEval: A Structured Benchmark for Evaluation of Vision-Language Models on Thermal Imagery}


\author{Ayush Shrivastava}
\email{shrivastavaayush@iitgn.ac.in}
\affiliation{%
  \institution{Indian Institute of Technology, Gandhinagar, India}
  % \city{Gandhinagar}
  % \state{Gujrat}
  \country{}
}

\author{Kirtan Ganagani}
\authornote{Both authors contributed equally to this research.}
\author{Laksh Jain}
\authornotemark[1]
\email{kirtan.gangani@iitgn.ac.in,laksh.jain@iitgn.ac.in}
% \email{}
\affiliation{%
  \institution{Indian Institute of Technology, Gandhinagar, India}
  % \city{Gandhinagar}
  % \state{Gujrat}
  \country{}
}

\author{Mayank Goel}
\email{mayankgoel@cmu.edu}
\affiliation{%
  \institution{Carnegie Mellon University, Pittsburgh, USA }
  % \city{Pittsburgh}
  \country{}
}

\author{Nipun Batra}
\email{nipun.batra@iitgn.ac.in}
\affiliation{%
  \institution{Indian Institute of Technology, Gandhinagar, India}
  % \city{Gandhinagar}
  % \state{Gujrat}
  \country{}
}

% \renewcommand{\shortauthors}{Ayush et al.}

\begin{abstract}
Vision language models (VLMs) achieve strong performance on RGB imagery, but they do not generalize to thermal images. Thermal sensing plays a critical role in settings where visible light fails, including nighttime surveillance, search and rescue, autonomous driving, and medical screening. Unlike RGB imagery, thermal images encode physical temperature rather than color or texture, requiring perceptual and reasoning capabilities that existing RGB-centric benchmarks do not evaluate. We introduce \textbf{ThermEval-B}, a structured benchmark of approximately \textbf{55{,}000} thermal visual question answering pairs designed to assess the foundational primitives required for thermal vision language understanding. ThermEval-B integrates public datasets with our newly collected \textbf{ThermEval-D}, the first dataset to provide dense per-pixel temperature maps with semantic body-part annotations across diverse indoor and outdoor environments. Evaluating 25 open-source and closed-source VLMs, we find that models consistently fail at temperature-grounded reasoning, degrade under colormap transformations, and default to language priors or fixed responses, with only marginal gains from prompting or supervised fine-tuning. These results demonstrate that thermal understanding requires dedicated evaluation beyond RGB-centric assumptions, positioning ThermEval as a benchmark to drive progress in thermal vision language modeling.

\end{abstract}






\maketitle

\section{Introduction}

\begin{figure*}
    \centering
    \includegraphics[width=1.0\linewidth]{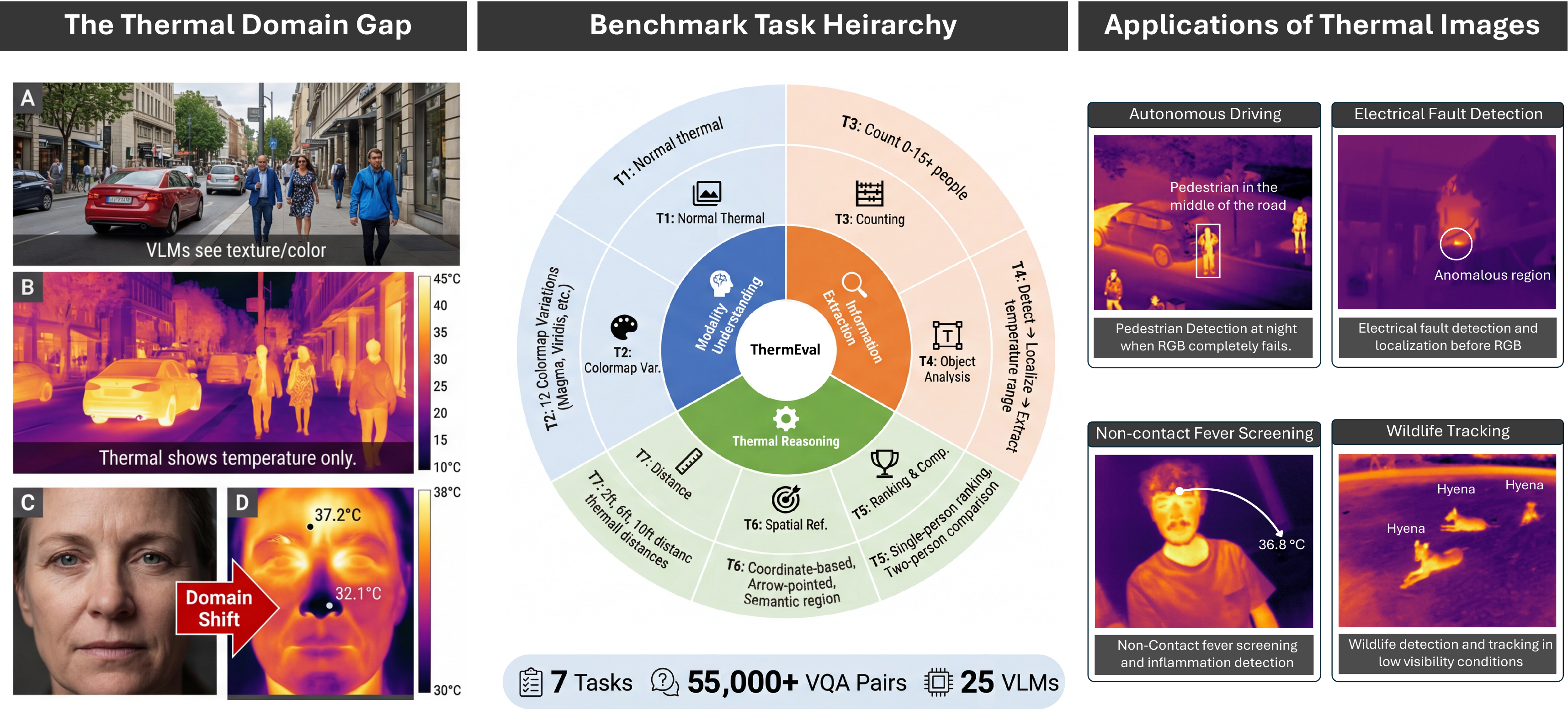}
    \caption{Thermal imagery enables critical perception tasks in settings where RGB fails, but VLMs trained predominantly on RGB exhibit systematic errors when applied to thermal images, driven by modality mismatch and language priors.}
    \label{fig:HeroFig}
\end{figure*}

Most computer vision research has traditionally focused on RGB images, which captures visible light with rich color and texture information. In contrast, thermal images measures emitted radiation and encodes temperature. This makes it fundamentally different from RGB images and valuable in conditions where visible light fails. As a result, thermal images lack many of the visual cues that standard vision models rely on, requiring alternative perceptual and reasoning capabilities. Thermal imagery plays a critical role in real world applications such as night time surveillance~\cite{jia2021llvip,El_Ahmar_2025_CVPR,liu2021cross}, search and rescue~\cite{peng_DroneCrowdcount,dronessurvey}, pedestrian detection, autonomous driving~\cite{flir2024dataset,hwang2015multispectral,Ashqar_MLLMs}, industrial fault detection~\cite{gasdetection,jadin_electrical}, and non contact medical applications~\cite{resprate_cvprw,nosebreath_cvprw,akbarian_jmir20,akbarian_jmir21}, where reasoning about temperature patterns is often more important than visual appearance. A brief visualization of these applications is shown in \blue{Figure}~\ref{fig:HeroFig}. For VLMs, interpreting thermal imagery therefore requires not only recognizing objects, but also grounding language in physically meaningful temperature signals. VLMs achieve strong performance on RGB, even without task-specific training, as evidenced by numerous benchmarks evaluating their capabilities across diverse RGB-centric tasks~\cite{roberts2025zerobench,fu2025mme,yu2023mmvet,li2024seed,li2024naturalbench}. However, it is not clear whether these models can generalize to thermal imagery. This leads to an important question: \textbf{Can VLMs trained predominantly on RGB data reason effectively about temperature specific tasks in thermal images?} The lack of benchmarks on thermal understanding makes it difficult to study this question in a systematic way.

\begin{figure*}
    \centering
    \includegraphics[width=1.0\linewidth]{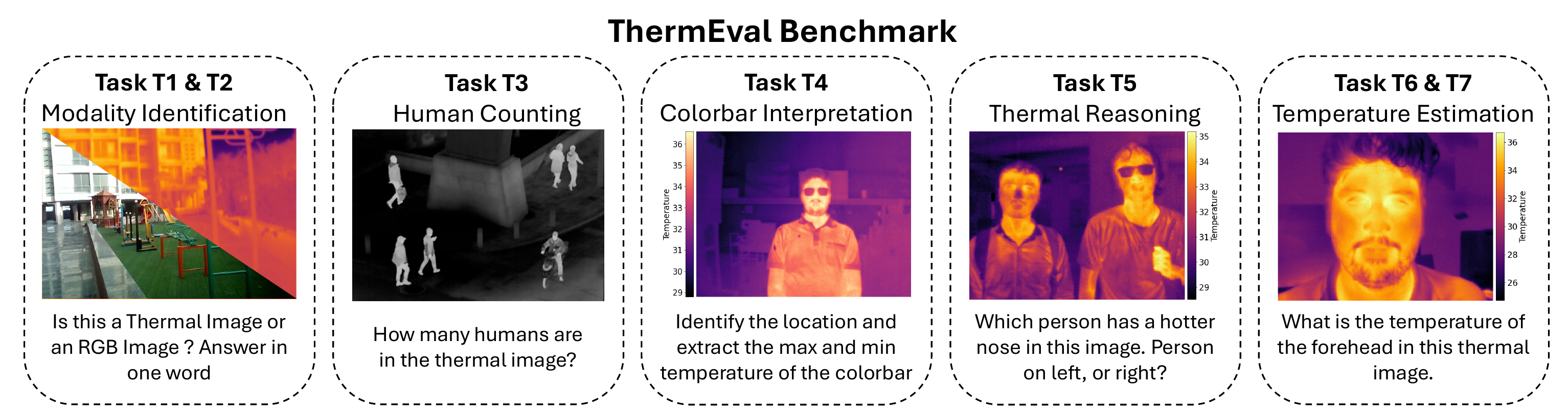}
    \vspace{-10pt}
    \caption{ThermEval defines seven evaluation tasks covering modality identification (T1–T2), human counting (T3), colorbar interpretation (T4), thermal reasoning (T5), and temperature estimation (T6–T7), designed to probe complementary aspects of thermal vision language understanding.}
    \label{fig:ThermEval_tasks}
\end{figure*}

To address this gap, we introduce ThermEval, a framework for evaluating VLMs on thermal imagery, consisting of a benchmark and an accompanying dataset. The benchmark ThermEval-B comprises seven tasks designed to reflect both fundamental challenges and real world applications (see \blue{Figure}~\ref{fig:HeroFig}~\&~\ref{fig:ThermEval_tasks} for an overview). These tasks cover modality identification (T1), robustness to colormap changes (T2), human counting (T3), colorbar interpretation (T4), thermal reasoning (T5), absolute temperature estimation (T6), and temperature interpretation at multiple depths (T7). By organizing tasks in increasing order of difficulty and targeting complementary aspects of thermal understanding, the benchmark encourages models to attend to thermal signals.

We evaluate 25 VLMs ranging from 0.3B to over 200B parameters, spanning both open-source and closed-source families, under standard zero-shot prompting~\cite{Brown_ZeroShot} and task-specific prompts. To estimate an upper bound on performance, we additionally fine-tune a representative model. While most VLMs reliably distinguish raw thermal from RGB images, performance degrades sharply on tasks requiring temperature reasoning or estimation. On reasoning tasks, models often rely on language priors rather than thermal cues, producing plausible but incorrect answers, such as defaulting to canonical body temperatures (36.8$^\circ$C) or fixed numeric outputs. Furthermore, models that fail at colorbar interpretation consistently underperform on temperature-grounded tasks, suggesting that errors on simpler prerequisite tasks predict failures on more complex thermal reasoning tasks. These failure modes appear across model scales, indicating that the limitation lies in cross-modal grounding rather than model size. Although fine-tuning improves performance to near-human levels on several tasks, models remain insufficiently reliable for real-world thermal applications. Overall, these results highlight thermal reasoning as an open challenge for current VLMs and underscore the need for dedicated benchmarks. This work makes the following contributions:

\begin{enumerate}[leftmargin=1.25em,nosep]
    \item We present \textbf{ThermEval-B}, a benchmark of approx \textbf{55,000 thermal VQA pairs} across seven tasks split between three datasets, providing the first systematic evaluation of VLMs on thermal imagery and revealing critical gaps in thermal reasoning.
    \item We introduce \textbf{ThermEval-D}, a dataset of over thousand thermal images with per-pixel temperature maps and body-part annotations across indoor and outdoor scenarios, supporting around \textbf{16k thermal VQA pairs} and enabling more realistic and comprehensive benchmarking than prior datasets.
\end{enumerate}

\section{\textbf{Related Work}} 

VLMs have demonstrated strong performance on RGB imagery, supported by benchmarks such as MME \cite{fu2025mme}, MMBench \cite{liu2024mmbench}, SEED-Bench\cite{li2024seed,li2024seed2}, and MMVet~\cite{yu2023mmvet,yu2024mmvet2}, which evaluate perception, reasoning, and problem-solving across diverse domains. More recent benchmarks, including NaturalBench~\cite{li2024naturalbench} and ZeroBench~\cite{roberts2025zerobench}, further challenge VLMs with adversarial samples and complex reasoning tasks. Despite these advances, existing evaluations remain largely RGB-centric and do not assess performance on alternative sensing modalities. We organize related work into three categories. First, we review benchmarks developed for thermal and multispectral modalities. Second, we survey existing thermal and infrared datasets. Finally, we discuss common evaluation practices and standard modeling assumptions used in thermal imagery analysis.

\subsection{Thermal and Multi-Spectral Benchmarks}
\textit{Existing benchmarks provide limited evaluation of the reasoning required for thermal understanding.} Multispectral modalities capture different physical signals, with thermal imagery representing temperature or depth encoding geometry. Recent works have begun to explore VLMs for non-RGB data. Chung et al.~\cite{chung2025enhancedvisionlanguagemodelsdiverse} use GPT-4o to generate multiple-choice questions for multispectral images, including thermal imagery, but rely on a constrained MCQ format that limits generality and reasoning depth. RGB-Th-Bench~\cite{moshtaghi2025rgbth} investigates RGB-thermal pairs, but is restricted to binary classification tasks and does not evaluate temperature interpretation; moreover, its evaluation protocol penalizes partially correct open-ended responses. In contrast, ThermEval targets thermal-specific challenges through structured tasks that require temperature reasoning and numerical estimation. Unlike prior binary or multiple-choice setups, our benchmark includes both classification and regression tasks with quantitative metrics, enabling a more faithful and fine-grained assessment of VLM performance on thermal imagery.

\subsection{Thermal and Infrared Datasets} 
\textit{Existing datasets provide limited support for temperature grounded reasoning due to the lack of dense, semantically meaningful temperature annotations.} Several public datasets are available, but only a small subset provides access to temperature values. Widely used datasets such as FLIR\_ADAS~\cite{flir2024dataset}, LLVIP~\cite{jia2021llvip}, SpeakingFaces~\cite{abdrakhmanova2021speakingfaces}, and TFW ~\cite{kuzdeuov2022tfw}  have advanced multimodal perception research, but they lack the pixel-level temperature annotations required for precise thermal reasoning. A limited number of datasets, including Charlotte-Faces~\cite{ashrafi2022charlotte} and the L-CAS Thermal Physiological Monitoring dataset~\cite{cosar2018thermal}, provide per-pixel temperature measurements, though these are restricted to facial imagery. The L-CAS RGBD-T dataset~\cite{cosar2019human} offers multimodal data but focuses primarily on human re-identification and does not include meaningful body-part annotations. As summarized in \blue{Table}~\ref{tab:dataset_comparison}, no existing dataset combines raw thermal imagery, per-pixel temperature maps, and diverse semantic contexts, a gap that ThermEval-D fills.

\begin{table}[t]
\centering
\caption{Comparison of thermal datasets, summarizing temperature data (Temp), bounding boxes (BBoxes), segmentation masks (Seg), annotator reliability (Rel), subject counts (Subj), and primary research objectives.}
\vspace{-1.0em}
\label{tab:dataset_comparison}
\resizebox{1.05\linewidth}{!}{
\setlength{\tabcolsep}{3pt} % Adjusted column spacing
\begin{tabular}{lcccccl}
\toprule
\textbf{Dataset} & \textbf{Temp} & \textbf{BBox} & \textbf{Seg} & \textbf{Rel} & \textbf{Subj} & \textbf{Objective} \\
\midrule
Charlotte~\cite{ashrafi2022charlotte} & \checkmark & \xmark & \xmark & \xmark & 10 & Facial thermography \\
M3FD~\cite{liu2022m3fd} & \xmark & \checkmark & \xmark & \xmark & many & Multi-modal object detection \\
LCAS Physio.~\cite{cosar2018thermal} & \checkmark & \xmark & \xmark & \xmark & 5 & Physiological monitoring \\
LCAS RGB-D-T~\cite{cosar2019human} & \checkmark & \checkmark & \xmark & \xmark & 15 & Human re-identification \\
FLIR ADAS~\cite{flir2024dataset} & \xmark & \checkmark & \xmark & \xmark & many  & ADAS object detection \\
LLVIP~\cite{jia2021llvip} & \xmark & \checkmark & \xmark & \xmark & many & Low-light pedestrian detection \\
SpeakingFaces~\cite{abdrakhmanova2021speakingfaces} & \xmark & \checkmark  & \xmark & \xmark & 142 & Speech and lipreading \\
TFW ~\cite{kuzdeuov2022tfw} & \xmark & \checkmark & \xmark & \xmark & 51 & Face and landmark recognition \\
\rowcolor{baseline}
\textbf{ThermEval (Ours)} & \checkmark & \checkmark & \checkmark & \checkmark & 35  & Vision Model Benchmarking \\
\bottomrule
\end{tabular}
}
\vspace{-1.25em}
\end{table}

\subsection{False-Colored Thermal Images}

\textit{Public datasets release thermal images in visualized form rather than as raw radiance or temperature matrices}, shaping how thermal imagery is used in practice. Raw sensor measurements are rarely available, and major datasets~\cite{ashfaq2021thermal,kniaz2018thermalgan,hwang2015multispectral,openthermalpose2023,jia2021llvip,flir2024dataset} instead provide false-colored thermal images, which have become the de facto representation for downstream thermal analysis. Our evaluation protocol follows established modeling practice, where physical sensor measurements are first visualized before being provided to the model. Prior work shows that VLMs can learn effectively from such visualized physical modalities, including thermal and depth data~\cite{spatialbot2025icra,mdpi2024thermalml,irgtp2025iccv,batterythermal2025eusipco,cai2025depthlmmetricdepthvision}. While VLMs are predominantly trained on RGB imagery, they are not inherently restricted to it, and false-colored thermal images provide a practical and widely adopted interface for evaluating multimodal reasoning.

\section{\textbf{ThermEval}}

We evaluate VLMs on thermal imagery using standardized prompts and a unified evaluation protocol to ensure consistent and reproducible results across models. All experiments are conducted under controlled settings without task-specific fine tuning unless explicitly stated. Additional details on the evaluation procedure and implementation are provided in \blue{Appendix}~\ref{B5}. The working code used for all experiments and analyses is publicly available in the  \href{https://github.com/AyushShrivstava/ThermEval_KDD}{\textcolor{blue}{github repository}} \footnote{\url{https://github.com/AyushShrivstava/ThermEval_KDD}} and \href{https://sustainability-lab.github.io/thermeval/}{\textcolor{blue}{project page}}\footnote{\url{https://sustainability-lab.github.io/thermeval/}}, where \blue{Appendix}~\ref{B2} describe it.

\subsection{ThermEval-B: Benchmark} \label{sec:Benchmark-tasks}

ThermEval-B is designed as a benchmark that evaluates increasingly complex capabilities required for thermal understanding . We begin by testing whether VLMs can recognize thermal imagery as a distinct modality (T1), which is a necessary prerequisite for any thermal reasoning. Since thermal images are commonly rendered using application dependent false colormaps, we next examine whether models rely on superficial color cues or can robustly identify thermal images under diverse colormap transformations (T2). Once basic modality awareness is established, we evaluate a standard semantic task, human presence and counting (T3), to assess whether models can extract meaningful content from thermal images. Before probing temperature estimation or reasoning, models must be able to localize the information needed for such tasks, motivating colorbar detection and interpretation (T4). We then assess relative thermal reasoning (T5), testing whether models can compare temperatures across and within individuals rather than relying on language based priors. Finally, we evaluate absolute temperature estimation (T6) and its robustness to imaging distance (T7), which together probe whether VLMs can extract and reason over numerical temperature values in realistic thermal sensing scenarios. Complete prompt templates used in our experiments are reported in \blue{Appendix}~\ref{B5}.

\subsection{Benchmark Tasks}

\subsubsection{\textbf{Modality Identification (T1 and T2):}}
This category evaluates whether VLMs can recognize thermal imagery based on its visual characteristics. In \textbf{T1}, we frame the problem as a modality classification task using paired thermal and RGB images from the FLIR and LLVIP datasets, with a balanced distribution of both modalities. \textbf{T2} extends this setting by testing robustness under colormap transformations, where thermal images are colorized using different colormaps commonly employed in practice. Although such colormaps preserve the underlying thermal signal, they alter visual appearance and may confuse models. We evaluate performance across sequential colormaps (Type I, e.g., Magma and Viridis) and more complex colormaps (Type II, e.g., Summer and Spring), relative to standard grayscale representations~\cite{matplotlib_colormaps}. The dataset remains identical to T1, with colormap transformations applied while retaining thermal modality as ground truth.

\subsubsection{\textbf{Human Presence and Counting (T3) :}}
This task evaluates a fundamental perceptual capability of VLMs: counting people in thermal images. We use thermal images from the FLIR and LLVIP datasets containing varying numbers of pedestrians in road scenes. Ground truth counts are derived from the person annotations provided with each dataset.

\subsubsection{\textbf{Inferring the Colorbar (T4) :}}
This task assesses whether VLMs can interpret the colorbar in thermal images, which is a prerequisite for temperature estimation and thermal reasoning. It consists of three subtasks: (1) colorbar detection, where the colorbar is absent in 50\% of images; (2) colorbar localization, identifying its position (Top, Left, Bottom, Right); and (3) temperature range extraction, requiring interpretation of the numerical scale. Ground truth is programmatically generated by placing colorbars at controlled locations with known ranges.

\subsubsection{\textbf{Thermal Reasoning (T5):}}
This task evaluates ability to reason about relative temperatures. It includes two subtasks: (1) comparative reasoning across individuals, where images contain two people and models compare body-part temperatures (chest, forehead, nose); and (2) within-individual reasoning, where models produce an ordered list of body parts for a person based on thermal intensity. Ground truth is obtained from human annotations, using mean body-part temperatures to determine the correct ordering.

\subsubsection{\textbf{Temperature Extraction (T6 and T7) :}}
Here we evaluate VLMs’ ability to estimate absolute temperatures from thermal images. This task is motivated from Meta's prior work (DepthLM~\cite{cai2025depthlmmetricdepthvision}) on estimating depth using VLMs. \textbf{T6} includes three difficulty levels: (1) coordinate-based estimation, where models extract temperatures at specified pixel coordinates annotated with visible axes (2) pixel-based estimation, where temperatures are queried at visually marked locations such as arrows/circle and (3) region-based estimation, where models infer temperatures for semantically meaningful regions. \textbf{T7} further examines robustness to imaging distance by prompting models to estimate temperatures of semantic regions at distances of 2 ft, 6 ft, and 10 ft. Ground truth for coordinate and pixel-based tasks is obtained programmatically, while region-based estimates use human annotations with mean body-part temperatures as reference values.

\subsection{\textbf{ThermEval-D: Dataset}}

We present ThermEval-D, the first thermal dataset covering both indoor and outdoor human-centric scenes with dense per-pixel temperature annotations. FLIR captures urban roads, LLVIP provides elevated street views, and ThermEval-D adds 1000 images from everyday environments such as offices, parks, and workspaces, each with detailed body-region annotations (forehead, chest, nose) enabling fine-grained tasks not possible using other datasets.
Task-wise VQA counts are provided in \blue{Table}~\ref{tab:task_vqa_counts_source}. Dataset is available on \href{https://www.kaggle.com/datasets/shriayush/thermeval}{\textcolor{blue}{kaggle}}\footnote{\url{https://www.kaggle.com/datasets/shriayush/thermeval}}.

\subsubsection{\textbf{Data Collection Protocol}}

The study was approved by the Institutional Ethics Committee (IEC), and all data were collected with informed consent and anonymized prior to release. We collected ThermEval-D across diverse indoor and outdoor environments within our institute, including offices, laboratories, workspaces, parks, and open grounds. Thirty-five adult participants (age 18–47, weight 64–108 kg) with varied skin tones participated voluntarily with written consent. Participants performed natural activities such as standing, sitting, walking, and navigating stairs, allowing us to capture varied postures and thermal profiles. Additional details are provided in \blue{Appendix}~\ref{A1}.

\subsubsection{\textbf{Dataset Annotation 
Details}}\label{sec:annotation-details}

Each thermal image in our dataset includes dense per-pixel temperature annotations, enabling fine-grained reasoning over spatial temperature patterns. Three expert annotators created polygonal segmentations of body parts following standardized guidelines with illustrative examples. Each image was annotated by all three annotators, and uncertainties were discussed collectively to ensure consistency across tasks. Semantic regions were defined as follows:

\noindent\textbf{Person:} Encompasses the entire visible human body, including limbs, while excluding accessories.    

\noindent\textbf{Forehead:} Extends from the hairline to the eyebrows, tightly cropped to avoid inclusion of eyes.   

\noindent\textbf{Nose:} From bridge to nostrils, excluding adjacent facial regions; glasses were excluded unless thermally indistinguishable.  

\noindent\textbf{Chest:} From base of neck to waistline, including shoulders and upper torso, excluding arms.  

\noindent Bounding boxes were automatically derived from polygons for compatibility across tasks, supporting both coarse and fine spatial resolutions. Inter-annotator agreement was high as measured by IoU and Dice metrics (BBox IoU 0.77, Segm. IoU 0.72, BBox Dice 0.87, Segm. Dice 0.84); pairwise results are summarized in \blue{Table}~\ref{tab:inter_annotator_agreement}.
.

\section{\textbf{Evaluation}}
In this section, we detail the VLMs used in our experiments and outline the evaluation protocol followed throughout the study.

\subsection{\textbf{Model Specifications}}

We evaluate 25 VLMs, including 17 open source models, 4 closed source models, 3 chart focused models, and one specialized fine tuned variant. We include chart focused VLMs because thermal images frequently contain embedded colorbars and numeric scales, requiring models to interpret visual legends and map colors to quantitative values. These models therefore provide a natural comparison point for temperature extraction and color grounded reasoning in thermal imagery. The evaluated models span a wide range of architectures and parameter scales, and selection is based on popularity and strong performance on established benchmarks. Our evaluation includes InternVL 3 (8B, 14B, 38B)~\cite{chen2024internvl}, LLaVA 1.5 (7B)~\cite{llava2024cot}, LLaMA 3.2 (11B)~\cite{llama3.2vision2024}, MiniCPM-V 2.6 (8B)~\cite{yao2024minicpm}, Phi-3 (4.2B), and Phi-3.5 (7B)~\cite{abdin2024phi3}. We also evaluate Qwen-VL (7B), Qwen-VL 2.5 (7B, 33B), Qwen A22 (235B)~\cite{bai2023qwenvl}, PaliGemma-2 (3B)~\cite{steiner2024paligemma2}, IDEFICS-3 (6.7B)~\cite{laurençon22024Idefics}, Smol-VLM (256M)~\cite{marafioti2025smolvlm}, Jina-VLM~\cite{koukounas2025jinavlm}, and BLIP-2 (9B)~\cite{li2023blip2}. The closed source group consists of Gemini 3 Pro, 3 Flash, 2.5 Pro, and 2.5 Flash~\cite{team2023gemini,team2024gemini}. We additionally evaluate chart focused models, namely ChartGemma (3B)~\cite{masry2024chartgemmavisualinstructiontuningchart}, ChartInstruct LLaMA-2 (7B)~\cite{masry2024chartinstruct}, and TinyCharts (3B)~\cite{zhang2024tinychart}. Finally, we include a fine tuned version of Qwen-VL 2.5 (7B) to assess the impact of domain specific adaptation on thermal reasoning tasks.

\subsection{\textbf{Evaluation Protocol}}
We evaluate all models under zero-shot prompting setting using fixed prompt templates and without any fine-tuning on thermal data. To further probe the potential for performance improvement through prompting alone, we conduct a systematic study on InternVL (14B), MiniCPM (8B), Qwen-VL-2.5 (7B), and BLIP-2 (9B), comparing standard zero-shot prompts with context-augmented variants. Finally, to assess the upper bound of achievable performance without architectural modifications, we perform supervised fine-tuning (SFT) of Qwen-VL-2.5 (7B) and evaluate the gains. 

\subsubsection{\textbf{LLM as a Parser:}}
Despite explicit instructions to produce fixed-format answers, VLM outputs often vary in structure (Figure~\ref{fig:HeroFig}). Although tools such as vLLM, Outlines, and Instructor~\cite{vllm,outlines,liu2024instructor} support structured output generation, they do not cover all models in our benchmark. Following prior work~\cite{danish2024geobenchvlm,zheng2023judging,gu2024survey}, we therefore use a language-only LLM (Gemini~2.5 models) to parse and standardize VLM outputs. The parser operates solely on the textual response, without access to the input image, and is invoked only when outputs deviate from the expected format, extracting class labels for classification tasks and numerical values for regression tasks. This approach enables consistent evaluation across heterogeneous outputs while using the LLM strictly as a parser rather than a scorer. Regex-based parsing proved brittle, whereas LLM-based parsing provided more robust extraction, consistent with prior observations~\cite{gu2024survey}.

\subsubsection{\textbf{Benchmarking the Parser :}} We validated our evaluation pipeline using a stratified gold set spanning all tasks and models, with correctness verified by human annotators. Structured judging was performed using Gemini~2.5 Pro, Gemini~2.5 Flash, and Gemini~2.5 Flash Lite through the Instructor framework, achieving agreement levels of 99.01\%, 99.07\%, and 98.24\%, respectively. Most remaining errors arose from ambiguous VLM responses rather than judge failures. The gold-set size was determined using standard statistical methods to ensure representativeness at the 95\% confidence level with a margin of error below 3\%. Additional details are provided in \blue{Appendix}~\ref{B4}.

\section{\textbf{Results}}

\subsection{Task 1 and Task 2: Modality Identification}  

Tasks 1 and 2 evaluate modality identification, with Task 2 introducing colormap transformations as a robustness challenge, as summarized in \blue{Table}~\ref{tab:tasks_1_4}. Human performance remains near perfect, with occasional errors attributable to annotation noise. In Task 1, most VLMs perform strongly, with Intern-VL 3 and Qwen-VL variants achieving near human accuracy, indicating that distinguishing RGB from raw thermal imagery is relatively straightforward. Notably, BLIP-2 consistently predicts all inputs as thermal images, which leads to poor performance on the Tasks. LLaVA-1.5 exhibits a similar failure mode, with a pronounced drop in performance from Task 1 to Task 2. Performance further varies across colormap types, where sequential colormaps such as Magma and Viridis are handled more reliably, while complex colormaps such as Summer and Spring induce larger failures. This pattern suggests that several models rely on low level color statistics rather than modality invariant representations. Overall, while VLMs handle basic modality identification well, their robustness to colormap transformations remains inconsistent, making Task 2 a stronger diagnostic of genuine thermal modality understanding. Detailed colormap specific results are provided in \blue{Appendix} \blue{Table}~\ref{tab:T2-appendix}.

\subsection{Task 3: Human Counting}
Task 3 evaluates VLMs’ ability to detect human presence and accurately count humans in thermal images. Results (\blue{Table}~\ref{tab:tasks_1_4}) reveal wide variability across models. Early-generation systems such as BLIP-2, PaliGemma-2, Phi and Idefics perform poorly, with MAE exceeding 4.0 on FLIR dataset. In contrast, more recent models like LLaMA-3.2, variants of Qwen-VL and InternVL, achieve substantial improvements, reducing error to around 3 on FLIR and less than 1 on LLVIP.  Scaling trends are evident within the Intern-VL family: the 8B model struggles (MAE $>$ 3 on FLIR), while the 14B and 38B variants improve markedly, with the 38B model reaching 2.93 on FLIR and 0.48 on LLVIP. Notably, Intern-VL (8B) exhibits a systematic failure, often defaulting to 11 when unable to resolve counts. Human annotators remain the most accurate, with MAE of 1.73 on FLIR and 0.3 on LLVIP.  \textit{Errors are most pronounced when images contain many individuals or overlapping thermal signatures, while both models and humans perform near-perfectly when counts are low or people are well separated.}. This persistent gap, especially on FLIR, highlights the difficulty of robust human counting in thermal imagery and underscores it as a key open challenge for VLM.

\begin{table*}[h]
\centering
\caption{VLM performance on Task-1 (modality identification), Task-2 (identification under colormap), Task-3 (human counting), and Task-4 (colorbar interpretation). Task 1,2 and 3 are evaulated using FLIR and LLVIP dataset while ThermEval was used for Task-4. $\uparrow$ indicates higher accuracy (ACC) is better. $\downarrow$ indicates lower mean absolute error (MAE) is better.}
\vspace{-1.0em}
\begin{adjustbox}{max width=\textwidth}
\renewcommand{\arraystretch}{0.9}
\begin{tabular}{l r rr rr rr rr rr}
\toprule
\multirow{4}{*}{\textbf{Model}}
& \multirow{4}{*}{\makecell{\textbf{Params} \\ \textbf{(in B)}}}
& \multicolumn{2}{c}{\textbf{Task-1}}
& \multicolumn{2}{c}{\textbf{Task-2}}
& \multicolumn{2}{c}{\textbf{Task-3}}
& \multicolumn{4}{c}{\textbf{Task-4}} \\
\cmidrule(lr){3-4} \cmidrule(lr){5-6} \cmidrule(lr){7-8} \cmidrule(lr){9-12}
&   & \multirow{2}{*}{\textbf{FLIR}} & \multirow{2}{*}{\textbf{LLVIP}}& \multirow{2}{*}{\textbf{FLIR} }& \multirow{2}{*}{\textbf{LLVIP}  }& \multirow{2}{*}{\textbf{FLIR}  }& \multirow{2}{*}{\textbf{LLVIP}  }&  \multicolumn{4}{c}{\textbf{ThermEval}} \\ 

\cmidrule(lr){9-12}
 &  & & & & & & &
\textbf{Detect} & \textbf{Position} &  \textbf{Max}  &  \textbf{Min} \\
% \midrule
\cmidrule(lr){3-4} \cmidrule(lr){5-6} \cmidrule(lr){7-8} \cmidrule(lr){9-12}
& & \textbf{ACC $\uparrow$} & \textbf{ACC $\uparrow$} & \textbf{ACC $\uparrow$} & \textbf{ACC $\uparrow$} & \textbf{MAE $\downarrow$} & \textbf{MAE $\downarrow$} & \textbf{ACC $\uparrow$} & \textbf{ACC $\uparrow$} & \textbf{MAE $\downarrow$} & \textbf{MAE $\downarrow$} \\
\midrule
\rowcolor{chartmodels}
ChartGemma & 3.0 &  0.50 &  0.50 &   0.00 &  0.00  & 3.04 & 1.25 &  0.48 & 0.45 & 0.04 & 0.03 \\
\rowcolor{chartmodels}
TinyCharts & 3.0 & 0.50 &  0.50 &   0.00 &  0.00 & 4.72 & 2.99 &  0.5 &  0.14 &  68.44 & 24.75 \\
\rowcolor{chartmodels}
ChartInstruct & 7.0 & 0.50 & 0.50 &  0.00 &  0.01 & 4.48 & 2.36 &  0.5 &  0.25 &  162.08 &  74.37 \\ 
\midrule
\rowcolor{opensource}
SMOL-256M & 0.3 & 0.03 & 0.10 & 0.28 & 0.33 & 4.31 & 1.77 & 0.41 & 0.00 & 0.22 & 3.00 \\
\rowcolor{opensource}
SAM-3 & 0.9 & - & - & - & -& 2.07 & 0.66 & - & - & - & - \\
\rowcolor{opensource}
Jina-VLM & 2.0 & \textbf{0.99} & \textbf{1.00} & \textbf{1.00} & \textbf{1.00} & 3.82 & 0.62 & \textbf{1.00} & \textbf{0.99} & \textbf{0.00} & \textbf{0.00} \\
\rowcolor{opensource}
PaliGemma-2      & 3.0   & 0.86 & 0.92 &  0.98 &  \textbf{1.00} &  4.03 &  1.02 &  0.96 &  0.79 & \textbf{0.00} & 0.12 \\
\rowcolor{opensource}
Phi-3            & 4.0 & 0.67 & 0.88 & 0.79 & 0.86 & 3.59 & 1.22 & \textbf{1.00} & 0.96 & \textbf{0.00} & \textbf{0.00} \\
\rowcolor{opensource}
Phi-3.5          & 4.0  & 0.86 & 0.83 & 0.95 & 0.98 & 4.42 & 1.07 & \textbf{1.00} & 1.00 & \textbf{0.00} & \textbf{0.00} \\
\rowcolor{opensource}
LLaVA-1.5        & 7.0   & 0.68 & 0.63 & 0.44 & 0.29 & 3.39 & 1.17 & 0.74 & 0.43 &  4.79 & 5.88 \\
\rowcolor{opensource}
IDEFICS-3        & 8.0 & 0.98 & 0.86 & 0.89 & 0.88 & 4.01 & 1.14 & 0.88 & 0.98 & 0.01 & 0.42 \\
\rowcolor{opensource}
Qwen-VL 2         & 8.0   & 0.98 & 0.98 & 0.98 & 0.99 & 3.66 & 0.79 & \textbf{1.00} & 0.95 & \textbf{0.00} & 0.01 \\
\rowcolor{opensource}
Qwen-VL 2.5      & 8.0   & \textbf{0.99} & \textbf{1.00} & \textbf{0.99} & \textbf{1.00} & 3.55 & 0.89 & \textbf{1.00} & \textbf{1.00} & \textbf{0.00} & \textbf{0.00} \\
\rowcolor{opensource}
Intern-VL 3      & 8.0   & \textbf{1.00} & \textbf{1.00} & \textbf{1.00} & \textbf{1.00} & 3.02 & 0.64 & \textbf{1.00} & \textbf{1.00} &  9.15 &  0.82 \\
\rowcolor{opensource}
MiniCPM-V 2.6    & 8.0   & 0.96 & 0.96 & 0.97 & 0.98 & 3.88 & 0.99 & \textbf{1.00} & 1.00 & 2.19 & 79.78 \\
\rowcolor{opensource}
BLIP-2           & 8.0   & 0.43 & 0.53 & 0.93 & 0.98 & 4.69 & 2.99 & 0.50 & 0.25 &  - &  - \\
\rowcolor{opensource}
LLaMA-3.2        & 11.0  & \textbf{1.00} & 0.92 & \textbf{1.00} & 0.90 & 2.84 & 0.73 & \textbf{1.00} & 0.91 & \textbf{0.00} & \textbf{0.00} \\
\rowcolor{opensource}
Intern-VL 3      & 15.0  & 0.96 & 0.99 & 0.92 & 0.98 & 2.99 & 0.71 & \textbf{1.00} & \textbf{1.00} & \textbf{0.00} & \textbf{0.00} \\
\rowcolor{opensource}
Qwen-VL 2.5      & 33.0  & 0.96   & \textbf{0.99} & 0.96 & \textbf{1.00} & 3.59 & 0.94 & 0.98 & \textbf{1.00} & \textbf{0.00} &  \textbf{0.00} \\
\rowcolor{opensource}
Intern-VL 3      & 38.0  & \textbf{0.99} & \textbf{1.00} & \textbf{1.00} & \textbf{1.00} & 2.93 & \textbf{0.48} & \textbf{1.00} & \textbf{1.00} & \textbf{0.00} & \textbf{0.00} \\
\midrule
\rowcolor{finetuned}
Qwen-VL 2.5 (SFT) & 8.0 & \textbf{1.00} & \textbf{1.00} & \textbf{1.00} & \textbf{1.00} & 1.85 & 0.55 & \textbf{1.00} & \textbf{1.00} & \textbf{0.00} & \textbf{0.00} \\
\midrule
\rowcolor{baseline}
\textbf{Human} & -- & 0.97 & 0.98 & 0.98 & 0.99 & \textbf{1.73} & \textbf{0.30} & \textbf{1.00} & \textbf{1.00} & \textbf{0.00} & \textbf{0.00} \\
\midrule
\rowcolor{baseline}
Random Chance & -- & 0.50 & 0.50 & 0.50 & 0.50 & -- & -- & 0.50 & 0.25 & -- & --\\
\bottomrule
\end{tabular}
\end{adjustbox}
\label{tab:tasks_1_4}
\end{table*}

\subsection{Task 4: Colorbar Interpretation}

This task evaluates the ability of VLMs to interpret thermal image colorbars which is a critical prerequisite for downstream temperature estimation and thermal reasoning tasks. As shown in \blue{Table}~\ref{tab:tasks_1_4}, while most modern VLMs achieve near-perfect accuracy, significant failures persist in models such as PaliGemma 2, LLaVA 1.5, IDEFICS, BLIP 2, and InternVL 3 (8B). Specifically, LLaVA, InternVL 3 (8B), and MiniCPM struggled with OCR, failing to reliably extract temperature extrema from the colorbars. LLaVA exhibited a tendency to hallucinate nonsensical, enormously large values, while InternVL 3 (8B) frequently committed decimal-shift errors (e.g.,$33.5 \rightarrow 335$). MiniCPM occasionally introduced a systematic error, adding a constant offset of 200 to the actual values. Furthermore, PaliGemma 2, LLaVA 1.5, and BLIP 2 struggled even with basic localization subtasks. In contrast, human performance remained perfectly accurate across all metrics. These results highlight a critical gap: while basic detection is common, several models still fail to reliably interpret the numerical temperature scales essential for thermal analysis. (See \blue{Table}~\ref{tab:T4-appendix} for detailed results.)

\begin{table*}[t]
\centering
\setlength{\tabcolsep}{8pt}
\caption{VLM performance on \textbf{Task-5} (Thermal reasoning), \textbf{Task-6} (Temperature estimation), and \textbf{Task-7} (Temperature estimation over varying depth). ThermEval dataset was used in evaluation of all the tasks in this table. $\uparrow$ indicates higher accuracy is better. $\downarrow$ indicates lower MAE is better. \textbf{?} indicates that weights are unknown. }
\vspace{-1.0em}
\begin{adjustbox}{max width=\textwidth}
\renewcommand{\arraystretch}{0.9}
\begin{tabular}{l r rr rrr rrr}
\toprule
\multirow{4}{*}{\textbf{Model}}
& \multirow{4}{*}{\makecell{\textbf{Params} \\ \textbf{(in B)}}}
& \multicolumn{2}{c}{\textbf{Task-5}}
& \multicolumn{3}{c}{\textbf{Task-6}}
& \multicolumn{3}{c}{\textbf{Task-7}} \\
\cmidrule(lr){3-4} \cmidrule(lr){5-7} \cmidrule(lr){8-10}
% & & \multicolumn{8}{c}{\textbf{ThermEval}}\\
% \cmidrule(lr){3-10}
&   & \textbf{Double} & \textbf{Single}
& \textbf{Coords} & \textbf{Marker}  & \textbf{Region} 
& \textbf{2ft}  & \textbf{6ft}  & \textbf{10ft}  \\
\cmidrule(lr){3-4} \cmidrule(lr){5-7} \cmidrule(lr){8-10}
&   & \textbf{Acc} $\uparrow$ & \textbf{Acc} $\uparrow$
& \textbf{MAE} $\downarrow$ & \textbf{MAE} $\downarrow$ & \textbf{MAE} $\downarrow$
& \textbf{MAE} $\downarrow$ & \textbf{MAE} $\downarrow$ & \textbf{MAE} $\downarrow$ \\
\midrule
\rowcolor{chartmodels}
ChartGemma & 3.0 & 0.46 & 0.36 & 13.74 & 3.84 & 2.33 & 1.20 & 1.02 & 1.08 \\
\rowcolor{chartmodels}
TinyCharts & 3.0 & 0.39 & 0.36 & 6.68 & 5.66 & 5.03 & 3.51 & 3.40 & 3.53 \\  
\rowcolor{chartmodels}
ChartInstruct & 7.0 & 0.00 & 0.17 & 9.37 & 4.02 & 4.38 & 2.96 & 2.76 & 2.37 \\
\midrule
\rowcolor{opensource}
SMOL-256M & 0.3 & 0.41 & 0.33 & 11.26 & 5.74 & 3.37 & 2.40 & 2.67 & 2.45 \\
\rowcolor{opensource}
Jina-VLM  & 2.0 & 0.39 & 0.32 & 4.22 & 3.26 & 1.95 & 1.24 & 1.05 & 0.81 \\
\rowcolor{opensource}
PaliGemma-2         & 3.0 & 0.53 & 0.19 & 58.42  & 3.52  & 2.53  & 1.49  & 1.29  & 1.27  \\
\rowcolor{opensource}
Phi-3               & 4.0 & 0.56 & 0.22 & 3.49  & 3.92  & 2.26  & 1.19  & 1.14  & 1.25  \\
\rowcolor{opensource}
Phi-3.5             & 4.0 & 0.39 & 0.38 & 3.14  & 3.55  & 2.07  & 1.43  & 1.48  & 1.37  \\
\rowcolor{opensource}
LLaVA-1.5           & 7.0 & 0.41 & 0.35 & 6.86 & 3.59  & 2.74  & 3.88  & 4.64  & 5.52  \\
\rowcolor{opensource}
IDEFICS-3           & 8.0 & 0.39 & 0.18 & 5.72  & 4.35  & 2.53  & 1.91  & 1.44  & 1.16  \\
\rowcolor{opensource}
Qwen-VL 2           & 8.0 & 0.41 & 0.51 & 3.28  & 3.30  & 2.01  & 1.25  & 1.00  & 0.89  \\
\rowcolor{opensource}
Qwen-VL 2.5         & 8.0 & 0.44 & 0.32 & 3.21  & 2.88  & 2.14  & 1.26 & 0.93  & 0.87  \\
\rowcolor{opensource}
Intern-VL 3         & 8.0 & 0.48 & 0.53 & 31.92 & 16.13 & 7.36 & 1.23  & 1.01 & 0.76 \\
\rowcolor{opensource}
MiniCPM-V 2.6       & 8.0 & 0.41 & 0.35 & 9.16  & 10.32 & 6.29  & 2.77  & 4.45  & 8.56  \\
\rowcolor{opensource}
BLIP-2              & 8.0 & 0.39 & 0.02 & 31.33 & 31.34 & - & - & - & - \\
\rowcolor{opensource}
LLaMA-3.2           & 11.0 & 0.61 & 0.42 & 3.00  & 3.99  & 3.03  & 2.39  & 1.74  & 1.48  \\
\rowcolor{opensource}
Intern-VL 3         & 15.0 & 0.53 & 0.40 & 2.60  & 3.64  & 2.22  & 1.24  & 1.03  & 0.89  \\
\rowcolor{opensource}
Qwen-VL 2.5         & 33.0 & 0.37 & 0.35 & 3.57  & 3.83  & 1.94  & 0.98  & 0.89  & 0.89  \\
\rowcolor{opensource}
Intern-VL 3         & 38.0 & 0.48 & 0.41 & 2.97  & 3.63  & 1.51  & 0.89  & 0.97  & 0.98  \\
\rowcolor{opensource}
Qwen A22            & 235.0  & 0.54 & 0.34 & 3.59 & 3.72 & 2.20 & 1.23 & 1.23 & 1.40 \\
\midrule
\rowcolor{finetuned}
Qwen-VL 2.5 (SFT) & 8.0 & 0.58 & 0.56 & \textbf{1.58} & \textbf{1.55} & \textbf{1.03} & \textbf{0.53} & \textbf{0.49} & \textbf{0.61} \\
\midrule
\rowcolor{closedsource}
Gemini 2.5 Flash & ? & 0.52 & 0.57 & 4.34 & 4.64 & 3.45 & 1.74 & 2.36 & 2.83 \\
\rowcolor{closedsource}
Gemini 2.5 Pro & ? & 0.57 & \textbf{0.63} & 3.47 & 3.26 & 2.32 & 1.47 & 1.95 & 2.66 \\
\rowcolor{closedsource}
Gemini 3 Flash & ? & 0.55 & 0.51 & 1.96 & 2.04 & 1.65 & 1.19  & 1.00 & 1.21 \\
\rowcolor{closedsource}
Gemini 3 Pro & ? & 0.74 & 0.61 & 1.94 & 1.86 & 1.47  & 1.00 & 0.74 & 0.90 \\

% & 0.28 & 3.81 & 3.48 & 2.50 & 1.30 & 1.80 & 1.96  \\
% \rowcolor{closedsource}
% Claude Haiku 4.5 & ? & 0.28 & \textbf{0.60} & 4.28 & 4.45 & 2.47 & 1.37 & 1.57 & 1.90 \\
% % \rowcolor{closedsource}
% GPT-4o & ? & 0.46 & 0.34 & x & x & x & x & x & x \\
\midrule
\rowcolor{baseline}
\textbf{Human} & -- & \textbf{0.84} & 0.54 & -- & 2.73 & 2.04 & 1.23 & 1.20 & 1.22 \\
\midrule
\rowcolor{baseline}
Random Chance & -- & 0.50 & 0.167 & -- & -- & -- & -- & -- & --\\
\bottomrule
\end{tabular}
\end{adjustbox}
\label{tab:tasks_5_6_7}
\end{table*}

\subsection{Task 5: Thermal Reasoning}  

Task 5 evaluates the ability of VLMs to reason over thermal intensities (\blue{Table}~\ref{tab:tasks_5_6_7}). In the comparative reasoning setting involving two individuals, accuracies for open-source models range from 0.37 to 0.61, with LLaMA 3.2 performing best. While older versions of Gemini perform on par with open-source models, the new Gemini 3 Pro achieves an accuracy of 0.74. However, this remains significantly below the human benchmark of 0.84. Within-individual reasoning which requires ranking specific body regions by thermal intensity proves even more challenging. Most open-source models perform poorly. Only Qwen2.5-VL (8B) and InternVL 3 (8B) achieve competitive scores of 0.51 and 0.53, respectively, compared to 0.54 for humans. Notably, models such as PaliGemma 2 and BLIP-2 collapse entirely on this task. Furthermore, scaling does not appear to provide clear gains. For instance, InternVL 3 (38B) lags behind its 8B and 15B variants. While closed-source models are comparable to or slightly outperform humans, these overall results highlight a fundamental limitation. \textit{Thermal reasoning demands structured relational understanding rather than just larger model scales}, underscoring the need for architectural innovation over simple parameter growth.

\subsection{Task 6 and Task 7: Temperature Estimation} 
Task 6 evaluates absolute temperature estimation from thermal images and Task 7 estimates the effect of depth in estimating temperature under controlled lab settings. Performance remains challenging for coordinate and marker based tasks in Task-6. Large open source models such as InternVL 3 (38B) and Qwen-VL 2.5 (38B) achieve MAEs above 3.5$^\circ$C and fail to reach the human baseline of 2.73$^\circ$C. Smaller models like PaliGemma and BLIP-2 perform drastically worse with MAEs exceeding 30$^\circ$C. This indicates a fundamental inability to map pixels to temperature values. Models that struggled with colorbar interpretation including InternVL 3 (8B), MiniCPM, LLaVA, PaliGemma, and BLIP-2 continue to underperform in temperature estimation. Notably, MiniCPM and InternVL (8B) occasionally ignore thermal inputs entirely and rely on language priors to output fixed values such as 37$^\circ$C (humans body temperature) for region based tasks. While older Gemini versions struggle, newer versions including Gemini 3 Pro and Flash successfully map pixel coordinates to temperatures. Region based estimation is more tractable as InternVL 3 (38B) achieves 1.51$^\circ$C and surpasses human performance. These results reveal that \textit{VLMs can default to prior biases rather than grounding predictions in thermal signals.} This underscores the need for architectures designed to truly interpret thermal imagery. \blue{Table}~\ref{tab:T6-Appendix} contains detailed results.

\subsection{Prompt Ablation}

We analyze prompt ablations to assess whether improved prompting can mitigate failures in thermal understanding. Implementation details are provided in \blue{Appendix}~\ref{prompt_ablation}. Models with reasonable visual grounding show substantial gains on simple modality recognition tasks when contextual modality descriptions are added. For instance, Qwen-VL~2.5 improves from 0.71 to 0.96 on T1 (FLIR) and from 0.61 to 0.98 on T2 (FLIR), while InternVL-14B reaches near-perfect accuracy on T2 after prompt augmentation. These gains suggest that short textual cues help anchor the visual signal for basic modality identification rather than revealing architectural limitations. In contrast, prompt ablations yield limited or inconsistent effects on tasks requiring thermal reasoning or temperature estimation. Across T5–T7, improvements are small, absent, or negative, with some models degrading under prompt augmentation. Models with weak initial thermal–visual alignment, such as BLIP-2, often perform worse when additional context is provided. Overall, these results show that prompt engineering can correct superficial modality confusion but cannot compensate for missing thermal-domain grounding, highlighting the need for training signals beyond prompt-level interventions.

\begin{table*}[t]
\centering
\caption{Performance of models before and after contextual prompt ablations. $\downarrow$ indicates lower and $\uparrow$ indicates higher is better. Colored deltas indicate improvement (\good{green}) or degradation (\bad{red}) relative to the simple prompt.}
\label{tab:prompt_ablation}
\vspace{-1.0em}
\setlength{\tabcolsep}{1.5pt} 
\resizebox{\textwidth}{!}{%
\begin{tabular}{llrrrrrrrr}
\toprule
\textbf{Task} & \textbf{Subtask} &
\makecell{\textbf{InternVL}\\\textbf{(Simple)}} &
\makecell{\textbf{InternVL}\\\textbf{(Contextual)}} &
\makecell{\textbf{MiniCPM}\\\textbf{(Simple)}} &
\makecell{\textbf{MiniCPM}\\\textbf{(Contextual)}} &
\makecell{\textbf{Qwen2.5-VL}\\\textbf{(Simple)}} &
\makecell{\textbf{Qwen2.5-VL}\\\textbf{(Contextual)}} &
\makecell{\textbf{BLIP-2}\\\textbf{(Simple)}} &
\makecell{\textbf{BLIP-2}\\\textbf{(Contextual)}} \\
\midrule
T1 & FLIR (ACC $\uparrow$)    & 0.96 & 0.97 (\good{+0.01}) & 0.95 & 0.96 (\good{+0.01}) & 0.71 & 0.96 (\good{+0.25}) & 0.46 & 0.34 (\bad{-0.12}) \\
   & LLVIP (ACC $\uparrow$)   & 1.00 & 1.00 (+0.00) & 0.98 & 0.99 (\good{+0.01}) & 0.72 & 0.96 (\good{+0.24}) & 0.22 & 0.43 (\good{+0.21}) \\
\hline
T2 & FLIR (ACC $\uparrow$)    & 0.86 & 0.99 (\good{+0.13}) & 0.91 & 0.89 (\bad{-0.02}) & 0.61 & 0.98 (\good{+0.37}) & 0.77 & 0.67 (\bad{-0.10}) \\
   & LLVIP (ACC $\uparrow$)   & 0.97 & 1.00 (\good{+0.03}) & 0.93 & 0.95 (\good{+0.02}) & 0.80 & 0.99 (\good{+0.19}) & 0.77 & 0.77 (+0.00) \\
\hline
T3 & FLIR (MAE $\downarrow$)  & 2.70 & 2.53 (\good{-0.17}) & 3.70 & 2.90 (\good{-0.80}) & 3.78 & 3.20 (\good{-0.58}) & 4.69 & 4.65 (\good{-0.04}) \\
   & LLVIP (AME $\downarrow$) & 0.73 & 0.60 (\good{-0.13}) & 1.09 & 0.80 (\good{-0.29}) & 1.09 & 0.88 (\good{-0.21}) & 2.99 & 2.94 (\good{-0.05}) \\ \hline
T4 & Detect (ACC $\uparrow$)   & 1.00 & 1.00 (+0.00) & 1.00 & 1.00 (+0.00) & 1.00 & 1.00 (+0.00) & 0.50 & 0.50 (+0.00) \\
   & Position (ACC $\uparrow$) & 1.00 & 1.00 (+0.00) & 0.99 & 0.99 (+0.00) & 0.99 & 0.97 (\bad{-0.02}) & 0.25 & 0.25 (+0.00) \\
   & Min (MAE $\downarrow$)    & 0.00 & 0.00 (+0.00) & 0.00 & 0.00 (+0.00) & 2.66 & 1.36 (\good{-1.30}) & 42.58 & 25.62 (\good{-16.96}) \\
   & Max (MAE $\downarrow$)    & 0.00 & 0.00 (+0.00) & 0.00 & 0.00 (+0.00) & 0.00 & 0.00 (+0.00) & 209.80 & 20.80 (\good{-189.00}) \\ 
   \hline
T5 & Single (ACC $\uparrow$)   & 0.32 & 0.20 (\bad{-0.12}) & 0.27 & 0.26 (\bad{-0.01}) & 0.42 & 0.39 (\bad{-0.03}) & 0.16 & 0.42 (\good{+0.26}) \\
   & Double (ACC $\uparrow$)   & 0.51 & 0.50 (\bad{-0.01}) & 0.40 & 0.41 (\good{+0.01}) & 0.41 & 0.51 (\good{+0.10}) & 0.39 & 0.39 (+0.00) \\ 
\hline
T6 & Arrow (MAE $\downarrow$)      & 5.28 & 4.30 (\good{-0.98}) & 6.32 & 5.95 (\good{-0.37}) & 4.75 & 4.46 (\good{-0.29}) & 12.74 & 12.77 (\bad{+0.03}) \\
   & Coordinate (MAE $\downarrow$) & 3.48 & 4.12 (\bad{+0.64}) & 4.00 & 4.85 (\bad{+0.85}) & 3.65 & 4.22 (\bad{+0.57}) & 13.08 & 12.74 (\good{-0.34}) \\
   & Region (MAE $\downarrow$)     & 2.18 & 2.51 (\bad{+0.33}) & 4.23 & 3.29 (\good{-0.94}) & 2.91 & 3.22 (\bad{+0.31}) & 14.73 & 14.73 (+0.00) \\
\hline
T7 & 2ft (MAE $\downarrow$)  & 1.01 & 2.04 (\bad{+1.03}) & 2.15 & 2.45 (\bad{+0.30}) & 1.05 & 1.63 (\bad{+0.58}) & 16.96 & 16.96 (+0.00) \\
   & 6ft (MAE $\downarrow$)  & 1.12 & 2.26 (\bad{+1.14}) & 2.02 & 2.47 (\bad{+0.45}) & 1.00 & 1.28 (\bad{+0.28}) & 16.35 & 16.35 (+0.00) \\
   & 10ft (MAE $\downarrow$) & 1.70 & 2.53 (\bad{+0.83}) & 1.85 & 2.01 (\bad{+0.16}) & 1.00 & 1.23 (\bad{+0.23}) & 15.43 & 15.43 (+0.00) \\
\bottomrule
\end{tabular}
}
\end{table*}

\begin{table}
\centering
\caption{Comparison of finetuned Qwen-VL 2.5 with human performance and other models. 
$\downarrow$ indicates lower is better and $\uparrow$ indicates higher is better. ``--'' indicates not applicable. \good{Green} in Qwen $\boldsymbol{\Delta}$ column indicates improvement.}
\vspace{-1.0em}
\label{tab:qwen_finetune_comparison}
\setlength{\tabcolsep}{3.5pt} 
\renewcommand{\arraystretch}{0.95}
\resizebox{\linewidth}{!}{%
\begin{tabular}{llrrrrr}
\toprule
\textbf{Task} & \textbf{Subtask} &
\makecell{\textbf{Best Model}\\\textbf{(Zero-shot)}} &
\textbf{Human} &
\makecell{\textbf{Qwen 2.5}\\\textbf{Zero-shot}} &
\makecell{\textbf{Qwen 2.5}\\\textbf{Finetuned}} &
\makecell{\textbf{Qwen}\\$\boldsymbol{\Delta}$} \\
\midrule
T1 & FLIR (Acc $\uparrow$)         & 1.00 & 0.97 & 0.98 & \textbf{1.00} & \good{+0.02} \\
   & LLVIP (Acc $\uparrow$)        & 1.00 & 0.98 & 1.00 & \textbf{1.00} & 0.00 \\
\midrule
T2 & FLIR (Acc $\uparrow$)         & 1.00 & 0.98 & 0.99 & \textbf{1.00} & \good{+0.01} \\
   & LLVIP (Acc $\uparrow$)        & 1.00 & 0.98 & 1.00 & \textbf{1.00} & 0.00 \\
\midrule
T3 & FLIR (MAE $\downarrow$)       & 2.72 & \textbf{1.73} & 3.55 & 1.85 & \good{$-1.70$} \\
   & LLVIP (MAE $\downarrow$)      & 0.51 & \textbf{0.30} & 0.89 & 0.55 & \good{$-0.34$} \\
\midrule
T4 & Detect (Acc $\uparrow$)       & 1.00 & \textbf{1.00} & 1.00 & \textbf{1.00} & 0.00 \\
   & Position (Acc $\uparrow$)     & 1.00 & \textbf{1.00} & 1.00 & \textbf{1.00} & 0.00 \\
   & Max (MAE $\downarrow$)        & 0.00 & \textbf{0.00} & 0.00 & \textbf{0.00} & 0.00 \\
   & Min (MAE $\downarrow$)        & 0.00 & \textbf{0.00} & 0.00 & \textbf{0.00} & 0.00 \\
\midrule
T5 & Double (Acc $\uparrow$)       & 0.61 & \textbf{0.84} & 0.44 & 0.58 & \good{+0.14} \\
   & Single (Acc $\uparrow$)       & \textbf{0.60} & 0.54 & 0.32 & 0.56 & \good{+0.24} \\
\midrule
T6 & Cords (MAE $\downarrow$)      & 3.48 & --    & 3.21 & \textbf{1.58} & \good{$-1.63$} \\
   & Arrow (MAE $\downarrow$)      & 3.48 & 2.73 & 2.88 & \textbf{1.55} & \good{$-1.33$} \\
   & Region (MAE $\downarrow$)     & 1.76 & 2.04 & 2.14 & \textbf{1.03} & \good{$-1.11$} \\
\midrule
T7 & 2ft (MAE $\downarrow$)        & 1.01 & 1.23 & 1.27 & \textbf{0.53} & \good{$-0.74$} \\
   & 6ft (MAE $\downarrow$)        & 1.00 & 1.20 & 0.94 & \textbf{0.49} & \good{$-0.26$} \\
   & 10ft (MAE $\downarrow$)       & 1.00 & 1.22 & 0.79 & \textbf{0.61} & \good{$-0.43$} \\
\bottomrule
\end{tabular}
}
\end{table}

\subsection{Supervised Fine-Tuning }

Supervised fine-tuning of Qwen-VL~2.5 (8B) yields substantial gains across ThermEval-B, enabling the model to outperform all evaluated zero-shot VLMs, including the much larger Qwen A22 (235B), and to match or exceed human performance on several tasks. We select Qwen-VL~2.5 for fine-tuning as it demonstrates strong and consistent zero-shot performance among open-source models on ThermEval-B, is widely adopted and representative of contemporary VLMs, and operates at a scale that enables reproducible fine-tuning on modest GPU setups. These improvements indicate that current VLMs possess the latent capacity for thermal understanding but lack adequate domain grounding when trained predominantly on RGB data. At the same time, fine-tuning does not fully resolve thermal reasoning: absolute temperature estimates still deviate by 1--2~$^\circ$C in T6 and T7, and semantic comparison performance in T5 remains below human level, error margins that are unacceptable for safety-critical applications such as fever screening or industrial monitoring. Together, these findings show that ThermEval provides meaningful supervision for diagnosing and improving thermal reasoning, while also highlighting the need for future VLM pretraining to explicitly incorporate physical sensor modalities rather than relying solely on RGB-centric representations. For implementation details on the fine-tuning experiment, kindly refer to \blue{Appendix}~\ref{SFT}.

\section{\textbf{Limitations}}\label{sec:limitations}

We evaluate a broad but non-exhaustive set of VLMs due to computational and access constraints, and use an LLM as an automatic judge to standardize heterogeneous outputs, which may introduce occasional parsing errors despite careful validation. ThermEval focuses on foundational thermal understanding tasks rather than highly complex scenarios, as our results show that current VLMs already struggle with basic prerequisites such as temperature estimation, and temperature-grounded reasoning. While more challenging tasks could be devised, reliable benchmarking at this foundational level is a necessary first step. Finally, a comprehensive thermal benchmark would ideally require raw per-pixel temperature matrices. However such data are rarely available in public datasets, and reliance on false-colored renderings currently limits benchmarking fidelity. These limitations motivate future work on richer datasets, access to raw sensor measurements, and progressively more challenging benchmarks built on ThermEval’s primitives.

\section{Acknowledgements}
We acknowledge Google for Gemini Academic Program Award which enabled us to run Gemini models reported in this work.

\section{\textbf{Conclusion}}
We introduce ThermEval, a structured benchmark for evaluating VLMs on thermal imagery. Across seven increasingly difficult tasks, we show that while current VLMs can identify thermal modalities and localize visual elements (e.g., colorbars), they consistently fail at core thermal reasoning and absolute temperature estimation, regardless of model scale. This suggests the key limitation is weak grounding in thermal signals and overreliance on language priors, not model capacity. Supervised fine-tuning improves results but remains insufficient for safety-critical use. Overall, ThermEval establishes thermal understanding as an open challenge and provides a diagnostic framework to drive future multimodal models that better integrate physical sensing with visual reasoning.

\clearpage

\bibliographystyle{ACM-Reference-Format}
\bibliography{sample-base}

% \appendix

\clearpage

\input{appendix}

\end{document}

%% file: appendix.tex
\appendix

\section*{\Large\textbf{Appendix}}

\vspace{1.5em}

\section*{Gen AI Disclosure}
We used generative AI tools, including ChatGPT and Gemini, to assist with improving the clarity, organization, and readability of the manuscript. Some parts of Figure~\ref{fig:HeroFig} were generated using Gemini. Apart from this figure, generative AI tools were used solely for language editing and presentation purposes. All content generated with the assistance of these tools was carefully reviewed, verified, and edited by the authors. All scientific ideas, experimental design, analyses, results, and conclusions are entirely the work of the authors. The use of generative AI did not influence the research methodology, data collection, model development, or interpretation of results.

\section{Datasets}  \label{A-Appendix}

\subsection{ThermEval-D Dataset}\label{A1}
We release ThermEval-D, a thermal image dataset with dense per-pixel temperature annotations, designed for tasks requiring precise temperature ground truths. The dataset contains over 1000 images of human subjects, each annotated with detailed regions including the forehead, chest, nose, and full-body presence. All imagery was captured using the TOPDON TC001 Plus thermal camera, which features a 256×192 pixel infrared sensor, sub-40 mK thermal sensitivity, 25 Hz frame rate, and a temperature measurement range of –20$^\circ$C to 550$^\circ$C with ±1$^\circ$C accuracy.

\begin{figure}[t]
    \centering
    \includegraphics[width=1.0\linewidth]{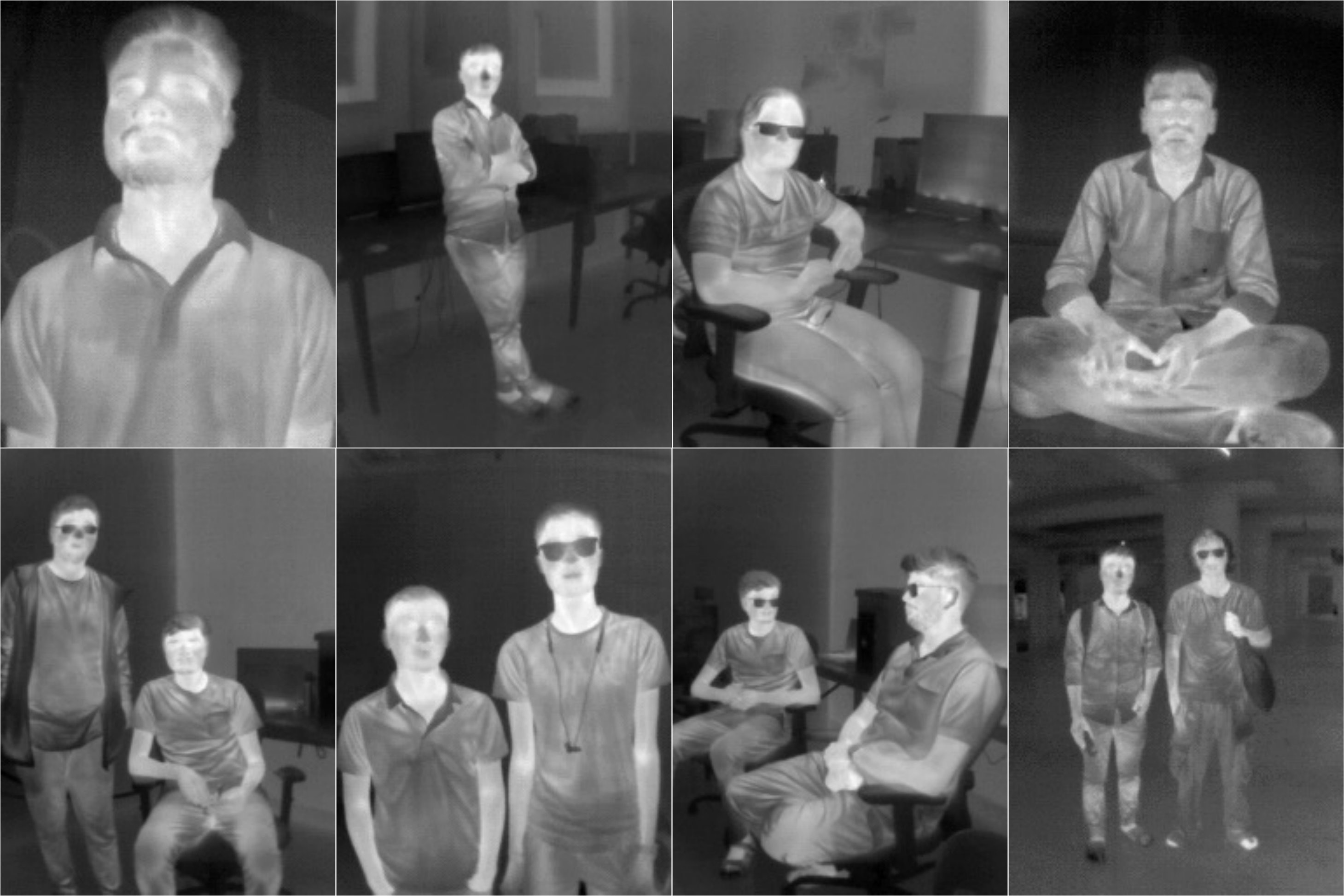}
    \caption{Images from ThermEval-D dataset. The top row shows the images having a single person in the scene whereas the second row shows the images having more than one person in the scene. Colorbars were added programatically during task evaluation}
    \label{fig:ThermEval_Dataset}
\end{figure}

ThermEval-D addresses the scarcity of thermal datasets with dense temperature data in the research community. The complete dataset, along with its accompanying croissant metadata file, is publicly accessible via  \href{https://www.kaggle.com/datasets/shriayush/thermeval}{\textcolor{blue}{Kaggle}}. A few sample of images from out Dataset are displayed in Figure~\ref{fig:ThermEval_Dataset}.

\textbf{Terms of Use and Licensing:} ThermEval-D is released under the Creative Commons Attribution-NonCommercial 4.0 (CC BY-NC 4.0) license, permitting unrestricted use for non-commercial research purposes.

\textbf{Data Maintenance and Accessibility:} The dataset is hosted on Kaggle, where we ensure long-term maintenance and periodic verification of accessibility. We plan regular expansions to enhance the dataset’s scope and utility for the research community. Our \href{https://github.com/AyushShrivstava/ThermEval_KDD}{\textcolor{blue}{benchmark}}
 involves ThermEval-D with other publicly available datasets for comprehensive evaluation across multiple tasks. While external datasets are used for comparative analysis, we do not redistribute them.

\textbf{ThermEval-D : Data Collection and Ethics:}
Data collection was conducted across diverse settings within the authors' institution, including parks, open grounds, offices, laboratories, and workspaces, following approval from the Institutional Ethics Committee (IEC). The dataset includes participants from various demographic groups, covering different genders, age ranges, body types, and heights, all performing distinct activities with informed consent. This study was approved by the IEC under the protocol titled “Thermal Image Benchmarking for VLMs,” valid from May 2025 for six months. All identifiable participant information was anonymized, and data collection posed minimal risk. Emergency medical support was readily available via the institutional medical center located approximately 100 meters from all collection sites.

\textbf{ThermEval-D Annotation Details :}
Each image was annotated by three expert annotators who created polygonal segmentations following standardized guidelines. Bounding boxes were automatically derived from these polygons to maintain compatibility across tasks and allow both coarse and fine spatial resolution. Inter-annotator agreement was quantified using pairwise IoU and Dice metrics for both bounding boxes and polygons, with mean values of 0.77 (BBox IoU), 0.72 (Segm. IoU), 0.87 (BBox Dice), and 0.84 (Segm. Dice), reflecting strong consistency; for context, even a one-pixel shift in a 10$\times$10 box yields IoU $\approx$ 0.68, confirming that observed values indicate true agreement rather than noise. Temperature variability across annotators was assessed by calculating the standard deviation of per-pixel temperatures within each segmentation, yielding a representative image example of 32.26$^\circ$C, 32.15$^\circ$C, and 32.18$^\circ$C (majority-vote 32.17$^\circ$C, std 0.04$^\circ$C), and a mean per-label standard deviation of 0.18$^\circ$C across the dataset, demonstrating robust and reliable temperature extraction. These procedures ensure that ThermEval-D provides accurate, consistent, and reproducible annotations for both spatial and temperature-based evaluation tasks.

\begin{table}[t]
\centering
\renewcommand{\arraystretch}{0.9}
\begin{tabular}{lcccc}
\toprule
\multirow{2}{*}{\textbf{Metric}} & \multicolumn{3}{c}{\textbf{Annotator Pairs}} & \multirow{2}{*}{\textbf{Mean of Pairs}} \\
\cmidrule(lr){2-4}
 & \textbf{1 \& 2} & \textbf{1 \& 3} & \textbf{2 \& 3} &  \\
\midrule
Bounding Box IoU  & 0.77 & 0.74 & 0.77 & 0.76 \\
Segmentation IoU  & 0.72 & 0.71 & 0.73 & 0.72 \\
Bounding Box Dice & 0.87 & 0.85 & 0.87 & 0.86 \\
Segmentation Dice & 0.84 & 0.83 & 0.84 & 0.84 \\
\bottomrule
\end{tabular}
\caption{Inter-annotator agreement using IoU and Dice}
\label{tab:inter_annotator_agreement}
\vspace{-1.5em}
\end{table}

\subsection{FLIR-ADAS Dataset}\label{A2}
\begin{figure}
    \centering
    \includegraphics[width=1.0\linewidth]{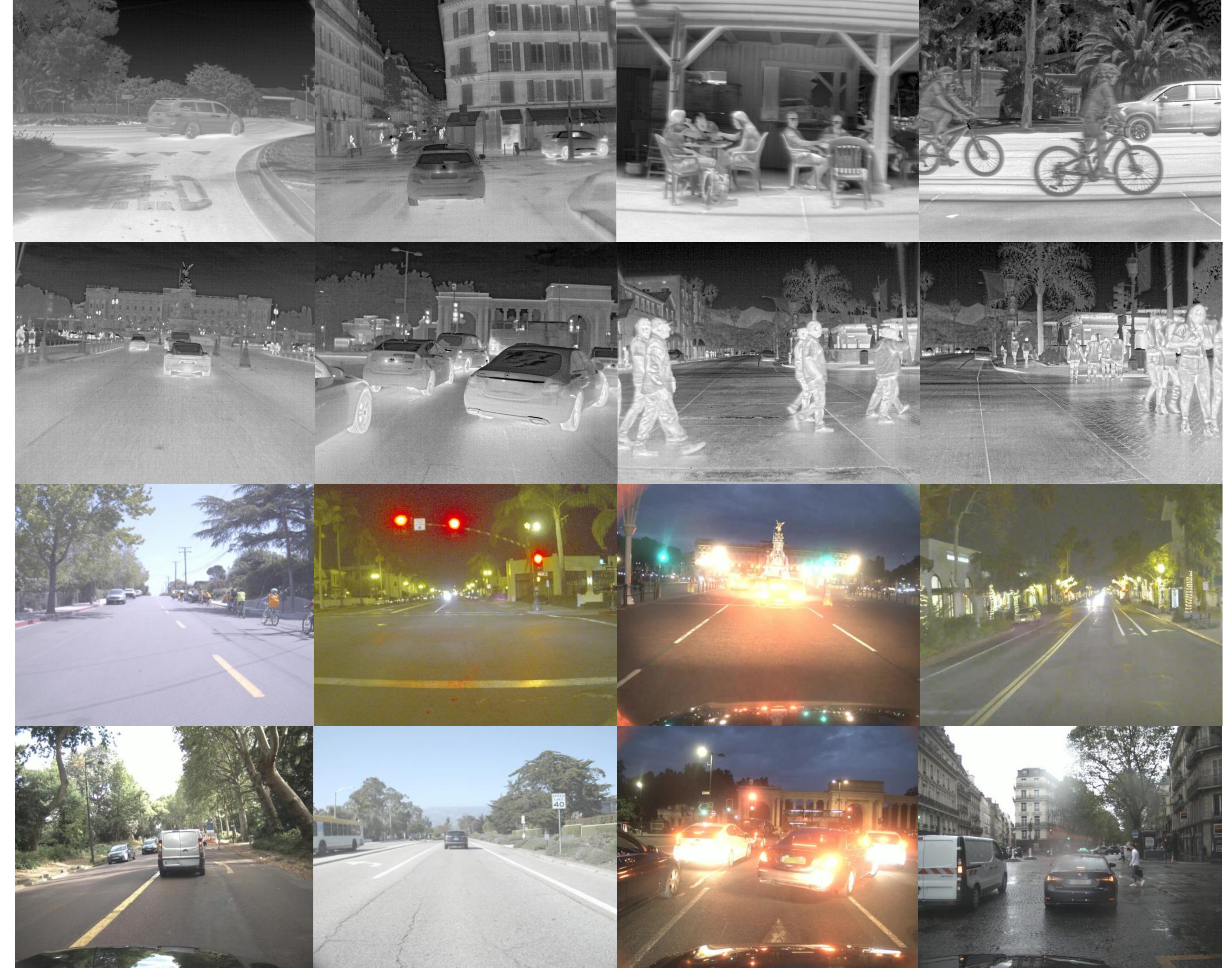}
    \captionof{figure}{Demonstrates some of the images from the FLIR-ADAS dataset, which is used for Tasks T-1, T-2, and T-3. Top row shows thermal images while the bottom shows RGB for different scenes. More information regarding the tasks could be obtained from section~\ref{sec:Benchmark-tasks}.}
    \label{fig:FLIR-ADAS}
\end{figure}

The FLIR-ADAS dataset\footnote{\url{https://adas-dataset-v2.flirconservator.com/\#downloadguide}} is a publicly available resource (separate from the ThermEval-D dataset release) designed to advance research in thermal-visible fusion (RGBT) algorithms for autonomous driving applications. This dataset contains approximately 13,000 aligned thermal and RGB image pairs with multi-class annotations, including pedestrian labels; however, it lacks temperature annotations. The thermal images maintain a consistent resolution of 640×512 pixels, while RGB image resolutions vary throughout the dataset. Samples from the FLIR dataset are illustrated in Figure~\ref{fig:FLIR-ADAS}.

\subsection{LLVIP Dataset}\label{A3}
The LLVIP dataset\footnote{\url{https://bupt-ai-cz.github.io/LLVIP/}} is also a publicly available dataset (not a part of the ThermEval-D dataset release) that has thermal and RGB aligned images aimed at advancing fusion techniques for pedestrian detection in low-light conditions. It consists of about 15000 thermal RGB image pairs annotated with people. Both thermal and RGB images maintain uniform 1280×1024 pixel resolution. Notably, this dataset lacks per-pixel temperature annotations for thermal imagery. Sample images from the LLVIP dataset are presented in Figure~\ref{fig:LLVIP}.

The FLIR-ADAS and LLVIP datasets were employed for Tasks T-1, T-2, and T-3, which evaluate fundamental VLM capabilities on thermal imagery without requiring specific temperature information. The ThermEval-D dataset was utilized for Tasks 4, 5 , 6 and 7, which necessitate precise temperature ground truth data for evaluation, a feature absent from existing publicly available thermal datasets.

\section{Implementation Details} \label{B-appendix}

\subsection{Compute Specifications}
\label{C1}
To ensure a fair comparison, all evaluations were conducted using the same hardware configuration: a single NVIDIA A100 GPU with 80GB of VRAM. Each evaluation involves a single forward pass (no ensembling or repeated sampling), and no access to model internals is assumed beyond what is publicly available through Hugging Face APIs or official released checkpoints. All prompt templates, prediction outputs, and evaluation scripts used in this study are provided in the accompanying GitHub repository.

\begin{figure}
    \includegraphics[width=0.985\linewidth]{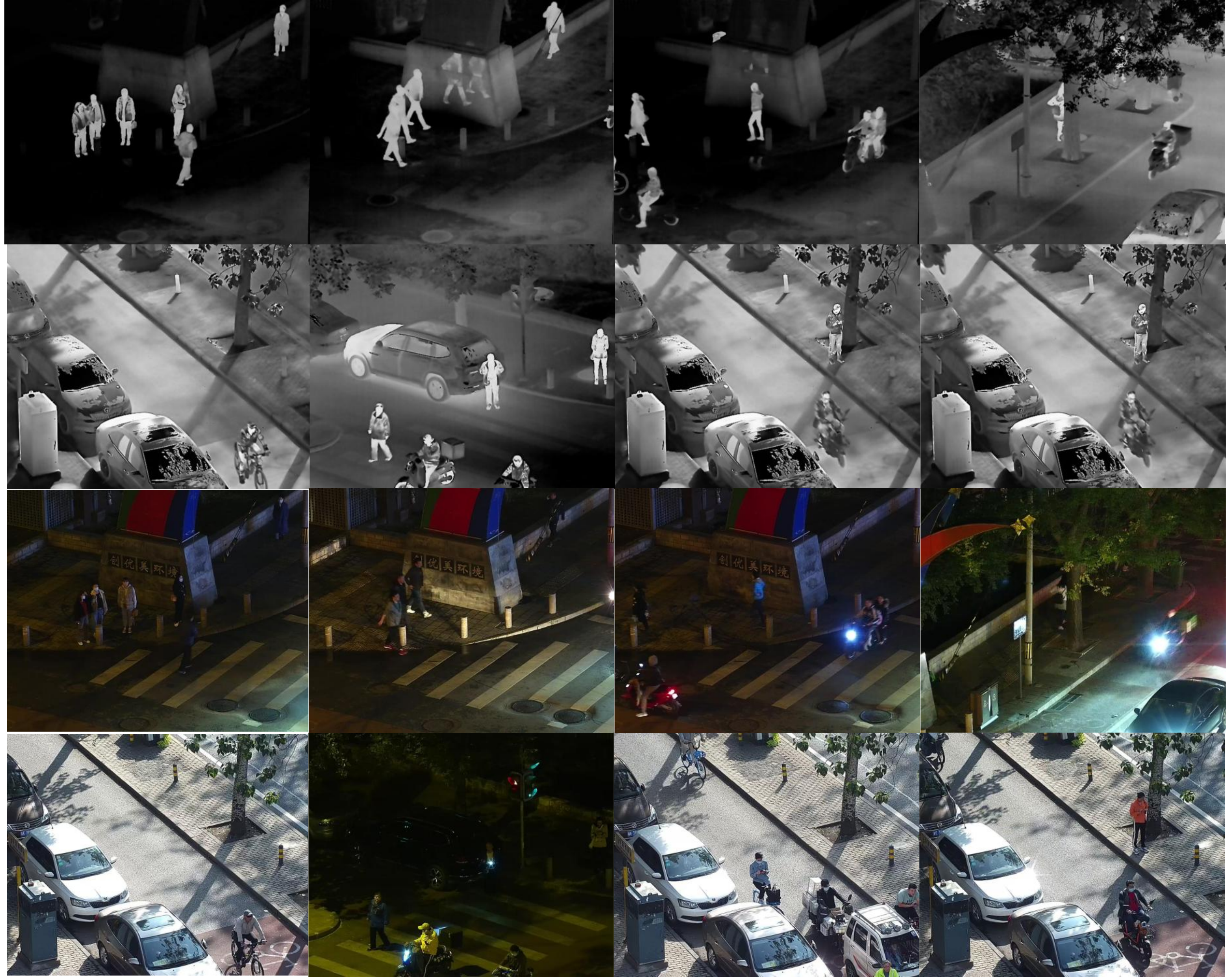}
    \captionof{figure}{Demonstrates some of the images from the LLVIP dataset, which is used for Tasks T-1, T-2, and T-3.Top row shows thermal images while the bottom shows RGB for different scenes. More information regarding the tasks could be obtained from section~\ref{sec:Benchmark-tasks}.}
    \label{fig:LLVIP}
\end{figure}

\subsection{Repository Structure}
\label{B2}
The repository is accessible \href{https://github.com/AyushShrivstava/ThermEval_KDD}{\textcolor{blue}{here}}. The root directory contains the following organizational structure:

1] \textbf{Dataset:} Contains all datasets utilized for model evaluation across different tasks.

2] \textbf{ThermEval-Benchmark:} Contains evaluation scripts for all tasks. These assess model performance across various tasks and saves the evalaution results, including the prompts used, correct answers and model outputs. Results are saved as a CSV separately for all the datasets.

3] \textbf{Labels:} Contains task-specific ground truth labels saved as CSV files for model evaluation. These files include image paths and corresponding ground truth such as modality, colourmap used, person count, temperature at given coordinates, etc, and other task-relevant annotations.

4] \textbf{Run.py:} The primary evaluation script for assessing vision-language models on all tasks. This script accepts model name as input parameter and saves evaluation results for the specified model in the evaluation results folder.
To evaluate additional models not specified in this paper, users need to define the corresponding load\_\{model\_name\} and infer\_\{model\_name\} functions in the model\_inference.py file located within the ThermEval\_Benchmark folder. Detailed instructions for this process are provided in the repository README.

\begin{table}[h]
\centering
\caption{VQA sample counts and data sources per task in ThermEval.}
\label{tab:task_vqa_counts_source}
\setlength{\tabcolsep}{7pt}
\resizebox{\columnwidth}{!}{%
\begin{tabular}{lll r}
\toprule
\textbf{ID} & \textbf{Task Name} & \textbf{Subtask / Source} & \textbf{\# Samples} \\
\midrule
T1 & Modality ID & FLIR, LLVIP & 10,000 \\
\midrule
T2 & \makecell[l]{Modality ID\\(Colormap)} & FLIR, LLVIP & 10,000 \\
\midrule
T3 & Counting & FLIR, LLVIP & 20,000 \\
\midrule
\multirow{4}{*}{T4} & \multirow{4}{*}{\makecell[l]{Colorbar\\Inference}} & Detection & 2,098 \\
 &  & Position & 4,196 \\
 &  & Maximum & 1,049 \\
 &  & Minimum & 1,049 \\
\midrule
\multirow{2}{*}{T5} & \multirow{2}{*}{\makecell[l]{Thermal\\Reasoning}} & Double & 155 \\
 &  & Single & 381 \\
\midrule
\multirow{3}{*}{T6} & \multirow{3}{*}{\makecell[l]{Temperature\\Estimation}} & Coords & 2,400 \\
 &  & Arrow & 2,400 \\
 &  & Region & 713 \\
\midrule
\multirow{3}{*}{T7} & \multirow{3}{*}{\makecell[l]{Temp Estimation\\(Distance)}} & 2 ft & 695 \\
 &  & 6 ft & 426 \\
 &  & 10 ft & 333 \\
\midrule
\multicolumn{3}{l}{\textbf{Total}} & \textbf{55,895} \\
\bottomrule
\end{tabular}
}
\end{table}

\subsection{\textbf{Model Evaluation Steps}}\label{B3}
\textbf{Setup:}

1] Download datasets: FLIR-ADAS (\href{https://adas-dataset-v2.flirconservator.com/#downloadguide}{\textcolor{blue}{link}}), LLVIP (\href{https://bupt-ai-cz.github.io/LLVIP/}{\textcolor{blue}{link}}), and ThermEval-D (\href{https://www.kaggle.com/datasets/shriayush/thermeval}{\textcolor{blue}{link}}) from provided links. 

2] Place the datasets (FLIR-ADAS, LLVIP and ThermEvalD) in the Datasets folder maintaining directory structure.

3] Create Python 3.8.10 virtual environment and install dependencies from requirements.txt.

\textbf{Execution:} 

4] Run Run.py from root directory, specify model name [for example: `qwen\_vl\_2\_5\_32B', `paligemma\_2\_3B', `internvl3\_38B', `llama\_3\_2\_11\_b', `minicpm\_2\_6']. Results are automatically saved to the Evaluation\_Results folder. Complete instructions are available in the repository README.

\subsection{\textbf{Sample Size Justification for LLM Parser Evaluation}} \label{B4}

To validate the LLM-based parser across all models and tasks, we created a gold set of approximately 1,200 parser outputs sampled from the full population of 700,000 outputs (50,000 VQA examples $\times$ 14 models). This sample size was chosen to provide statistically reliable estimates of parser accuracy while keeping annotation costs manageable.

Using the standard formula for finite-population proportions:

\[
n = \frac{Z^2 \, p \, (1-p)}{e^2} \cdot \frac{N}{N-1 + \frac{Z^2 \, p \, (1-p)}{e^2}}
\]

where $n$ is the required sample size, $N = 700{,}000$ is the population of outputs, $p = 0.5$ is the conservative estimate for expected parser accuracy, $e = 0.03$ is the desired margin of error, and $Z = 1.96$ corresponds to a 95\% confidence level, we obtain $n \approx 1,067$. This confirms that sampling approximately 1,200 outputs provides a 95\% confidence interval of $\pm 3\%$ for proportion-based metrics such as exact match accuracy.

To ensure the gold set is representative, we performed stratified random sampling across tasks, models, and answer types, including edge cases such as multi-number outputs and malformed answers. This approach guarantees coverage of the full distribution of parser outputs, allowing us to estimate parser performance accurately for the entire population of 700,000 VLM outputs.

\subsection{\textbf{Additional Results and Task-Wise Implementation Details}}
\label{B5}

\subsubsection{\textbf{Task 1: Modality Identification}}

\leavevmode

\noindent \textbf{Task: } This task aims to understand whether VLMs can visually distinguish RGB and Thermal Images.

\noindent \textbf{Zero-shot prompt:} Is this a thermal image or an RGB image? Strictly answer in one word.

\noindent \textbf{Implementation details:} This is a binary classification task, making its evaluation simple. We used 5,000 thermal-RGB image pairs each from the FLIR and LLVIP datasets, ensuring an equal number of thermal and RGB images for fair assessment. Sample images used for this task are shown in Figures~\ref{fig:FLIR-ADAS} and \ref{fig:LLVIP}.

\begin{table*}[t]
\centering
\setlength{\tabcolsep}{3pt} 
\small
\begin{tabularx}{\textwidth}{l c*{5}{>{\centering\arraybackslash}X >{\centering\arraybackslash}X}}
\toprule
\multirow{2}{*}{\textbf{Model}} & \multirow{2}{*}{\textbf{Params (B)}} & \multicolumn{2}{c}{\textbf{Gray}} & \multicolumn{2}{c}{\textbf{Magma}} & \multicolumn{2}{c}{\textbf{Spring}} & \multicolumn{2}{c}{\textbf{Summer}} & \multicolumn{2}{c}{\textbf{Viridis}} \\
\cmidrule(lr){3-4} \cmidrule(lr){5-6} \cmidrule(lr){7-8} \cmidrule(lr){9-10} \cmidrule(lr){11-12}
& & \textbf{FLIR} & \textbf{LLVIP} & \textbf{FLIR} & \textbf{LLVIP} & \textbf{FLIR} & \textbf{LLVIP} & \textbf{FLIR} & \textbf{LLVIP} & \textbf{FLIR} & \textbf{LLVIP} \\
\midrule

SMOL-256M & 0.3 &
0.002 & 0.003 & 0.634 & 0.553 & 0.268 & 0.575 & 0.250 & 0.241 & 0.519 & 0.456 \\
Jina-VLM & 2.0 &
1.000 & 1.000 & 1.000 & 1.000 & 1.000 & 1.000 & 1.000 & 1.000 & 1.000 & 1.000 \\
PaliGemma-2 & 3.0 &
0.995 & 0.997 & 1.000 & 1.000 & 0.992 & 1.000 & 0.915 & 1.000 & 1.000 & 1.000 \\
Phi-3 & 4.0 &
0.262 & 0.669 & 0.978 & 0.910 & 0.865 & 0.931 & 0.907 & 0.863 & 0.936 & 0.937 \\
Phi-3.5 & 4.0 &
0.751 & 0.899 & 1.000 & 0.992 & 1.000 & 1.000 & 0.995 & 1.000 & 1.000 & 0.998 \\
LLaVA-1.5 & 7.0 &
0.229 & 0.440 & 0.978 & 0.771 & 0.128 & 0.012 & 0.195 & 0.006 & 0.662 & 0.208 \\
IDEFICS -3 & 8.0 &
0.958 & 0.667 & 1.000 & 0.985 & 0.654 & 0.961 & 0.853 & 0.916 & 0.989 & 0.879 \\
Qwen-VL 2 & 8.0 &
0.961 & 0.954 & 1.000 & 1.000 & 0.971 & 1.000 & 0.975 & 0.996 & 1.000 & 1.000 \\
Qwen-VL 2.5 & 8.0 & 0.997 & 0.996 & 1.000 & 1.000 & 0.973 & 1.000 & 0.993 & 1.000 & 1.000 & 1.000 \\
Intern-VL 3 & 8.0 &
0.994 & 1.000 & 1.000 & 0.998 & 1.000 & 0.998 & 0.996 & 0.999 & 1.000 & 1.000 \\
MiniCPM-V 2.6 & 8.0 &
0.902 & 0.947 & 1.000 & 0.992 & 0.974 & 1.000 & 0.981 & 0.962 & 1.000 & 1.000 \\
BLIP-2 & 8.0 &
0.678 & 0.983 & 1.000 & 0.992 & 0.997 & 0.998 & 0.987 & 0.980 & 1.000 & 0.951 \\
LLaMA-3.2 & 11.0 &
0.999 & 0.722 & 1.000 & 1.000 & 0.994 & 0.941 & 0.995 & 0.835 & 1.000 & 0.995 \\
Intern-VL 3 & 15.0 &
0.899 & 0.996 & 0.993 & 0.972 & 0.971 & 1.000 & 0.804 & 0.972 & 0.909 & 0.964 \\
Qwen-VL 2.5 & 33.0 &
0.900 & 0.978 & 1.000 & 1.000 & 0.913 & 1.000 & 0.988 & 1.000 & 1.000 & 1.000 \\
Intern-VL 3 & 38.0 &
1.000 & 1.000 & 1.000 & 1.000 & 1.000 & 1.000 & 1.000 & 1.000 & 1.000 & 1.000 \\

\bottomrule
\end{tabularx}
\caption{Accuracy of VLMS on Task-2: Modality Identification under colormap transformation with results shown separately for FLIR and LLVIP datasets. Higher numbers are better.}
\label{tab:T2-appendix}
\end{table*}

\begin{table*}[t]
\centering
\setlength{\tabcolsep}{2pt} 
\small
\begin{tabularx}{\textwidth}{l r *{4}{>{\centering\arraybackslash}X} *{4}{>{\centering\arraybackslash}X}}
\toprule
\multirow{2}{*}{\textbf{Model}} & \multirow{2}{*}{\textbf{Params (B)}} 
& \multicolumn{4}{c}{\textbf{FLIR}} & \multicolumn{4}{c}{\textbf{LLVIP}} \\
\cmidrule(lr){3-6} \cmidrule(lr){7-10}
& & \textbf{MAE ↓} & \textbf{STD ↓} & \textbf{Bias *} & \textbf{RMSE ↓} 
& \textbf{MAE ↓} & \textbf{STD ↓} & \textbf{Bias *} & \textbf{RMSE ↓} \\
\midrule

SMOL-256M           & 0.3 & 4.31 & 5.56 & -4.30 & 7.03 & 1.77 & 1.73 & -1.63 & 2.38 \\
Jina-VLM              & 2.0  & 3.82 & 5.22 & -3.80 & 6.45 & 0.62 & 0.90 & -0.45 & 1.01 \\
PaliGemma-2 & 3.0 & 4.03 & 5.37 & -4.01 & 6.70 & 1.02 & 1.30 & -0.90 & 1.58 \\
Phi-3 & 4.0  & 3.59 & 4.96 & -3.52 & 6.08 & 1.22 & 1.23 & -1.17 & 1.70 \\
Phi-3.5 & 4.0   & 4.42 & 100.08 & -2.14 & 100.10 & 1.07 & 1.16 & -1.01 & 1.53 \\
LLaVA-1.5           & 7.0   & 3.39 & 5.28 & -3.08 & 6.11 & 1.17 & 1.60 & -0.52 & 1.69 \\
IDEFICS-3           & 8.0   & 4.01 & 5.25 & -4.00 & 6.60 & 1.14 & 1.21 & -1.04 & 1.59 \\
Qwen-VL 2           & 8.0   & 3.66 & 5.16 & -3.65 & 6.32 & 0.79 & 1.06 & -0.60 & 1.22 \\
Qwen-VL 2.5         & 8.0   & 3.55 & 5.06 & -3.51 & 6.16 & 0.89 & 1.10 & -0.80 & 1.36 \\
Intern-VL 3          & 8.0   & 3.02 & 4.58 & -2.98 & 5.46 & 0.64 & 1.01 & -0.21 & 1.03 \\
MiniCPM-V 2.6       & 8.0   & 3.88 & 5.29 & -3.86 & 6.55 & 0.99 & 1.22 & -0.88 & 1.50 \\
BLIP-2              & 8.0 & 4.69 & 5.59 & -4.69 & 7.30 & 2.99 & 1.82 & -2.99 & 3.50 \\
LLaMA-3.2           & 11.0  & 2.84 & 4.15 & -2.63 & 4.91 & 0.73 & 1.08 & -0.14 & 1.09 \\
Intern-VL 3          & 15.0  & 2.99 & 4.54 & -2.95 & 5.41 & 0.71 & 1.00 & -0.55 & 1.14 \\
Qwen-VL 2.5         & 33.0  & 3.59 & 5.11 & -3.52 & 6.20 & 0.94 & 1.21 & -0.69 & 1.39 \\
Intern-VL 3          & 38.0  & 2.93 & 4.51 & -2.91 & 5.36 & 0.48 & 0.78 & -0.27 & 0.83 \\

\bottomrule
\end{tabularx}
\caption{Regression metrics for Task-3 : Human Counting using FLIR and LLVIP datasets.  ↓ indicates lower is better. *Bias closer to 0 is better.}
\label{tab:T3-appendix}
\end{table*}

\subsubsection{\textbf{Task 2: Modality Identification under Colormap Transformations}}

\leavevmode

\noindent \textbf{Task:} This task extends task 1 by evaluating VLMS on thermal images with colormap transformations.

\noindent \textbf{Zero-shot prompt:} Is this a thermal image or an RGB image? Strictly answer in one word.

\begin{figure}[t]
    \centering
    \includegraphics[width=1.0\linewidth]{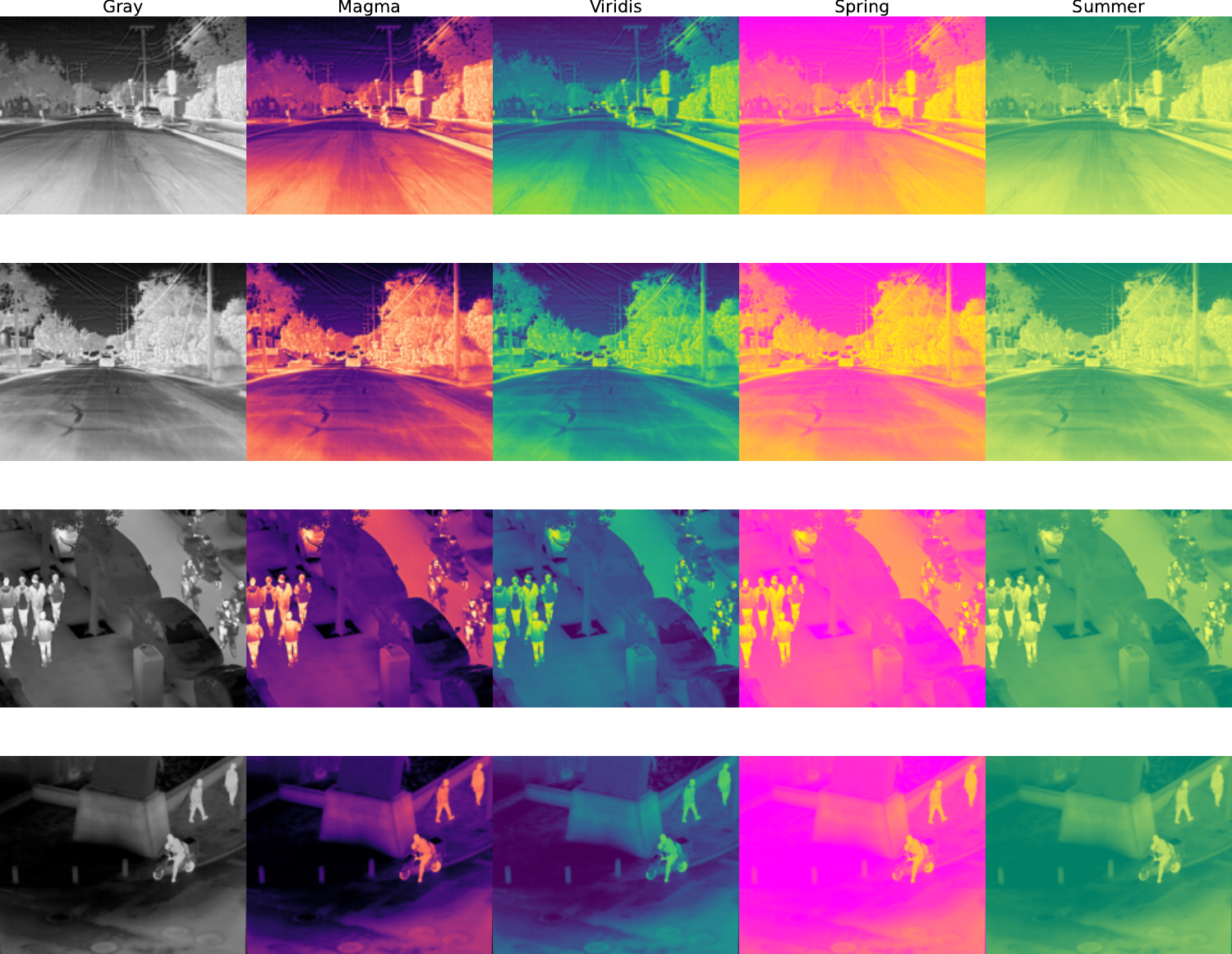}
    \caption{Demonstrates various colormaps used for Task T-2. Colormaps used were `gray', `magma', `viridis', `spring' and `summer'}
    \label{fig:colormap_illustration}
    % \vspace{-1.5em}
\end{figure}

\noindent \textbf{Implementation Details:} This is a binary classification task, making its evaluation simple. We used 1,000 thermal images each from the FLIR and LLVIP datasets, applying five colormap transformations per image to create a total of 10,000 images. We used simple sequential colormaps (Type I) such as Magma and Viridis, and more complex ones (Type II) like Summer and Spring, along with standard grayscale thermal images. Sample images used in this task are shown in Figures~\ref{fig:colormap_illustration}.

\begin{table*}[ht]
\centering
\setlength{\tabcolsep}{4pt} 
% \resizebox{\textwidth}{!}{%
\begin{tabular}{@{}l r r r rrrrrr@{}}
\toprule
\textbf{Model} & \textbf{Params (B)} & \textbf{Detection} & \textbf{Position} & \multicolumn{6}{c}{\textbf{Extraction}} \\
\cmidrule(lr){5-10}
& & Accuracy & Accuracy & Acc Max & Acc Min & Acc & MAE Max & MAE Min & MAE \\
\midrule

SMOL-256M & 0.3 & 0.414 & 0.000 & 0.990 & 0.824 & 0.907 & 0.22 & 3.0 & 1.61 \\
Jina-VLM & 2.0 & 1.00 & 0.994 & 1.000 & 1.000 & 1.000 & 0.0 & 0.0 & 0.0 \\
PaliGemma-2 & 3.0 & 0.961 & 0.785 & 1.000 & 0.955 & 0.978 & 0.0 & 0.12 & 0.06 \\
Phi-3 & 4.0 & 1.00 & 0.958 & 1.000 & 0.995 & 0.998 & 0.0 & 0.0 & 0.0 \\
Phi-3.5 & 4.0 & 1.00 & 1.00 & 0.999 & 1.000 & 1.000 & 0.0 & 0.0 & 0.0 \\
LLaVA-1.5 & 7.0 & 0.736 & 0.430 & 0.010 & 0.010 & 0.010 & 4.79 & 5.88 & 5.33 \\
IDEFICS-3 & 8.0 & 0.88 & 0.98 & 0.997 & 0.990 & 0.993 & 0.0 & 0.42 & 0.21 \\
Qwen-VL 2 & 8.0 & 1.00 & 0.954 & 1.000 & 0.979 & 0.990 & 0.0 & 0.01 & 0.0 \\
Qwen-VL 2.5 & 8.0 & 1.00 & 1.00 & 1.000 & 1.000 & 1.000 & 0.0 & 0.0 & 0.0 \\
Intern-VL 3 & 8.0 & 1.00 & 1.00 & 0.907 & 0.992 & 0.949 & 9.15 & 0.82 & 4.98 \\
MiniCPM-V 2.6 & 8.0 & 1.00 & 0.996 & 0.737 & 0.106 & 0.421 & 2.19 & 79.78 & 41.12 \\
BLIP-2 & 8.0 & 0.50 & 0.25 & -- & -- & -- & -- & -- & -- \\
LLaMA-3.2 & 11.0 & 1.00 & 0.913 & 1.000 & 1.000 & 1.000 & 0.0 & 0.0 & 0.0 \\
Intern-VL 3 & 15.0 & 1.00 & 1.00 & 1.000 & 1.000 & 1.000 & 0.0 & 0.0 & 0.0 \\
Qwen-VL 2.5 & 33.0 & 0.984 & 0.999 & 1.000 & 1.000 & 1.000 & 0.0 & 0.0 & 0.0 \\
Intern-VL 3 & 38.0 & 1.00 & 1.00 & 1.000 & 1.000 & 1.000 & 0.0 & 0.0 & 0.0 \\

\bottomrule
\end{tabular}%
% }
\caption{Model evaluation for Task-4: colorbar interpretation task, assessing the ability to detect, position, and extract temperature values. Acc Max and Acc Min denotes the accuracy of correctly identifying maximum and minimum values of the colorbar. MAE Max and MAE Min denotes the MAE is estimating Max and Min temperature of the colorbar.}
\label{tab:T4-appendix}
\end{table*}

\subsubsection{\textbf{Task 3: Counting Humans}}

\leavevmode

\noindent \textbf{Task:} This task assesses the basic object counting capability of VLMS, specifically focusing on counting people.

\noindent \textbf{Zero-shot prompt:} How many people are in this image? If there are no people, return 0. Stricly answer in integer.

\noindent \textbf{Implementation Details:} This regression task used 10,000 grayscale thermal images each from the FLIR and LLVIP datasets. A separate model parsed the outputs to estimate the numerical count of people.

\subsubsection{\textbf{Task 4: Reading Colorbar}}

\leavevmode

\noindent \textbf{Task:} This task evaluates the VLMs ability to identify and read the colorbar. It comprises of 3 subtasks (a) Identifying the presence of colorbar, (b) Identifying the location of the colorbar (top, left, bottom or right), and (c) Extracting the max and min value on the Colorbar.

\noindent \textbf{Zero-shot prompt subtask 1:}  You are given a thermal image. Does it contain a color bar or temperature scale that maps colors to temperature values? Answer only with 'Yes' or 'No'. Strictly answer in one word.

\noindent \textbf{Zero-shot prompt subtask 2:}  You are given a thermal image. It contains a color bar or temperature scale that maps colors to temperature value. What is the location of the colorbar? Possible locations are top, left, bottom, right. Strictly answer in one word.

\noindent \textbf{Zero-shot prompt  subtask 3:}  You are given a thermal image with a color bar or temperature scale that maps colors to temperature value. What is the {minimum / maxximum} temperature value in degree Celsius? Strictly return a single numerical value rounded to one decimal place?

\noindent \textbf{Implementation Details:} This task contains both classification as well as regression task. Prompt 1 and 2 would lead to a classification task where as the task 3 would lead to regression task. From subtask (a) the random chance accuracy is 50\% whereas for subtask (b) the random chance accuracy is 25\%.

\begin{table*}[t]
\centering
\setlength{\tabcolsep}{3pt} 
\resizebox{\textwidth}{!}{%
\begin{tabular}{@{}l r rrrr rrrr rrrr@{}}
\toprule
\textbf{Model} & \textbf{Params (B)} & \multicolumn{4}{c}{\textbf{Arrow}} & \multicolumn{4}{c}{\textbf{Coordinates}} & \multicolumn{4}{c}{\textbf{Region}} \\
\cmidrule(lr){3-6} \cmidrule(lr){7-10} \cmidrule(lr){11-14}
& & \multicolumn{1}{c}{\textbf{MAE}$\downarrow$} & \multicolumn{1}{c}{\textbf{RMSE}$\downarrow$} & \multicolumn{1}{c}{\textbf{BIAS}*} & \multicolumn{1}{c}{\textbf{STD}$\downarrow$} & \multicolumn{1}{c}{\textbf{MAE}$\downarrow$} & \multicolumn{1}{c}{\textbf{RMSE}$\downarrow$} & \multicolumn{1}{c}{\textbf{BIAS}*} & \multicolumn{1}{c}{\textbf{STD}$\downarrow$} & \multicolumn{1}{c}{\textbf{MAE}$\downarrow$} & \multicolumn{1}{c}{\textbf{RMSE}$\downarrow$} & \multicolumn{1}{c}{\textbf{BIAS}*} & \multicolumn{1}{c}{\textbf{STD}$\downarrow$} \\
\midrule

ChartGemma              & 3.0   & 3.84 & 5.05 & 1.19 & 4.91 &13.74 &28.06 & 8.43 &26.77 & 2.33 & 3.44 & -0.61 & 3.39 \\
TinyCharts              & 3.0   & 5.66 & 6.64 & -2.92 & 5.96 & 6.68 &16.62 & 1.32 &16.57 & 5.03 & 5.92 & -0.27 & 5.92 \\
ChartInstruct           & 7.0   & 3.96 & 5.94 & -1.47 & 5.75 & 9.37 &33.94 & 1.09 &33.93 & 4.38 & 6.30 & -0.81 & 6.25 \\

SMOL-256M               & 0.3   & 5.74 & 6.87 & 5.35 & 4.31 &11.26 &33.12 & 5.45 &32.67 & 3.37 & 4.55 &  1.95 & 4.11 \\
Jina-VLM                & 2.0   & 3.26 & 4.34 & 0.37 & 4.32 & 4.22 &14.95 & -1.12 &14.90 & 1.95 & 2.83 & -0.16 & 2.83 \\
PaliGemma-2             & 3.0   & 3.52 & 4.78 & 1.23 & 4.62 &58.42 &711.59 &53.05 &709.61 & 2.53 & 3.67 & -0.20 & 3.67 \\
Phi-3                   & 4.0   & 3.92 & 5.31 & 1.24 & 5.17 & 3.49 & 4.68 & -0.71 & 4.63 & 2.26 & 3.24 &  0.65 & 3.17 \\
Phi-3.5                 & 4.0   & 3.55 & 4.75 & 0.34 & 4.73 & 3.14 & 4.26 & -0.38 & 4.25 & 2.07 & 2.88 &  0.74 & 2.78 \\
LLaVA-1.5               & 7.0   & 3.59 & 4.67 & 2.45 & 3.98 & 6.86 &79.36 & 2.05 &79.33 & 2.74 & 3.56 &  2.34 & 2.69 \\
IDEFICS-3               & 8.0   & 4.35 & 6.44 & 1.41 & 6.29 & 5.72 &14.58 & 4.61 &13.84 & 2.53 & 3.93 & -0.42 & 3.91 \\
Qwen-VL 2               & 8.0   & 3.30 & 4.60 & 0.55 & 4.57 & 3.28 & 4.55 & -2.33 & 3.90 & 2.01 & 3.11 & -0.52 & 3.06 \\
Qwen-VL 2.5             & 8.0   & 2.88 & 3.84 & -0.03 & 3.84 & 3.21 & 4.35 & -1.15 & 4.19 & 2.14 & 2.80 &  0.15 & 2.80 \\
Intern-VL 3 (8B)        & 8.0   &16.13 &76.09 & -6.41 &75.82 &31.92 &90.79 &24.85 &87.33 & 7.36 &29.28 & -1.41 &29.24 \\
MiniCPM-2.6             & 8.0   &10.32 &20.72 & -4.58 &20.20 & 9.16 &17.26 & -3.82 &16.84 & 6.29 &13.90 & -4.54 &13.13 \\
BLIP-2                  & 8.0   &31.34 &31.52 &-31.34 & 3.35 &31.33 &31.52 &-31.18 & 4.60 &  -   &  -   &   -   &  -   \\
LLaMA-3.2               &11.0   & 3.99 & 5.34 & 2.04 & 4.93 & 3.00 & 4.27 &  1.02 & 4.15 & 3.03 & 4.35 & -1.41 & 4.12 \\
Intern-VL 3 (14B)       &15.0   & 3.64 & 4.87 & -0.62 & 4.83 & 2.60 & 3.38 &  0.22 & 3.37 & 2.22 & 3.04 & -1.23 & 2.78 \\
Qwen-VL 2.5 (32B)       &33.0   & 3.83 & 5.15 & -0.15 & 5.14 & 3.57 & 4.92 & -1.99 & 4.49 & 1.94 & 2.60 &  0.18 & 2.60 \\
Intern-VL 3 (38B)       &38.0   & 3.63 & 4.88 & -0.09 & 4.88 & 2.97 & 4.03 & -1.58 & 3.70 & 1.51 & 2.04 & -0.09 & 2.04 \\
Qwen A22                &235.0  & 3.72 & 5.01 & 1.52 & 4.77 & 3.59 & 4.75 &  1.44 & 4.53 & 2.20 & 3.27 &  1.37 & 2.97 \\

Gemini 2.5 Flash        & ?     & 4.64 & 5.94 & 1.37 & 5.78 & 4.34 & 5.54 &  0.45 & 5.52 & 3.45 & 4.45 &  3.00 & 3.29 \\
Gemini 2.5 Pro          & ?     & 3.26 & 4.62 & 0.64 & 4.58 & 3.47 & 4.66 &  1.15 & 4.52 & 2.32 & 2.79 &  1.81 & 2.12 \\
Gemini 3 Flash          & ?     & 2.04 & 3.00 & -0.21 & 2.99 & 1.96 & 2.85 &  0.32 & 2.83 & 1.65 & 2.12 &  1.08 & 1.83 \\
Gemini 3 Pro            & ?     & 1.86 & 2.77 & -0.53 & 2.72 & 1.94 & 2.62 & -0.99 & 2.43 & 1.47 & 1.91 &  0.04 & 1.91 \\
\bottomrule
\end{tabular}%
}
\caption{Regression metrics for Task-6: Temperature Estimation on ThermEval Dataset.  ↓ indicates lower is better. *Bias closer to 0 is better.}
\label{tab:T6-Appendix}
\end{table*}

\subsubsection{\textbf{Task 5: Temperature Reasoning}}

\leavevmode

\noindent \textbf{Task:} This task evaluates the reasoning capabilities of VLMS in thermal domain. It comprises of 2 subtasks: (a) Ranking the chest, head and nose of a person from hottest to coldest and (b) To compare the temperature head/chest/nose of 2 people in the image and return ``left" or ``right".

\noindent \textbf{Zero-shot prompt subtask 1:}  Given the thermal image and the colourbar, rank the following body parts in order from highest to lowest temperature: [bodyparts]. List them from hottest to coolest. Strictly return a list of body parts.

\noindent \textbf{Zero-shot prompt subtask 2:}  Given the thermal image with colourbar, determine whether the \{body\_part\} of the left person or the right person is hotter. Respond with only 'left' or 'right'. Strictly answer in one single word.

\noindent \textbf{Implementation Details:} This task involves binary classification and ordering, using the ThermEval-D dataset as it requires the temperature ground truths. The thermal image of size 256 x 192 which is same as the size of the temperature matrix, ie, 256 x 192, and mean temperatures were computed for regions defined by polygon box coordinates.

\begin{table*}[t]
\centering
\setlength{\tabcolsep}{3pt} 
\label{tab:T7-Appendix}
\resizebox{\textwidth}{!}{%
\begin{tabular}{l r rrrr rrrr rrrr}
\toprule
\textbf{Model} & \textbf{Params (B)} & \multicolumn{4}{c}{\textbf{2ft}} & \multicolumn{4}{c}{\textbf{6ft}} & \multicolumn{4}{c}{\textbf{10ft}} \\
\cmidrule(lr){3-6} \cmidrule(lr){7-10} \cmidrule(lr){11-14}
& & \textbf{MAE}$\downarrow$ & \textbf{RMSE}$\downarrow$ & \textbf{BIAS}* & \textbf{STD}$\downarrow$ & \textbf{MAE}$\downarrow$ & \textbf{RMSE}$\downarrow$ & \textbf{BIAS}* & \textbf{STD}$\downarrow$ & \textbf{MAE}$\downarrow$ & \textbf{RMSE}$\downarrow$ & \textbf{BIAS}* & \textbf{STD}$\downarrow$ \\
\midrule

ChartGemma & 3.0 & 1.20 & 1.49 & -0.66 & 1.33 & 1.02 & 1.28 & -0.30 & 1.25 & 1.08 & 1.36 & 0.06 & 1.36 \\
TinyCharts & 3.0 & 3.51 & 4.12 & 1.12 & 3.96 & 3.40 & 3.79 & 2.04 & 3.19 & 3.53 & 3.79 & 2.70 & 2.66 \\
ChartInstruct & 7.0 & 2.96 & 3.18 & -1.45 & 2.83 & 2.76 & 2.98 & -1.62 & 2.50 & 2.37 & 2.66 & -1.75 & 2.01 \\

SMOL-256M & 0.3 & 2.40 & 2.85 & 0.72 & 2.75 & 2.67 & 3.02 & 0.35 & 3.00 & 2.45 & 2.82 & 0.26 & 2.81 \\
Jina-VLM & 0.3 & 1.24 & 1.57 & -0.97 & 1.23 & 1.05 & 1.31 & -0.90 & 0.96 & 0.81 & 1.01 & -0.56 & 0.83 \\
PaliGemma-2 & 3.0 & 1.49 & 1.77 & -0.79 & 1.58 & 1.29 & 1.54 & -0.18 & 1.53 & 1.27 & 1.52 & 0.58 & 1.41 \\
Phi-3 & 4.0 & 1.19 & 1.48 & -0.09 & 1.48 & 1.14 & 1.38 & 0.07 & 1.38 & 1.25 & 1.48 & 0.54 & 1.38 \\
Phi-3.5 & 4.0 & 1.43 & 1.77 & -0.41 & 1.72 & 1.48 & 1.79 & -0.36 & 1.76 & 1.37 & 1.68 & -0.01 & 1.68 \\
LLaVA-1.5 & 7.0 & 3.88 & 4.22 & 3.88 & 1.68 & 4.64 & 4.94 & 4.61 & 1.77 & 5.52 & 5.75 & 5.52 & 1.62 \\

IDEFICS-3 & 8.0 & 1.91 & 2.27 & -0.75 & 2.14 & 1.44 & 1.72 & -0.62 & 1.60 & 1.16 & 1.49 & -0.49 & 1.40 \\
Qwen-VL 2 & 8.0 & 1.25 & 1.58 & -0.95 & 1.26 & 1.00 & 1.24 & -0.50 & 1.14 & 0.89 & 1.12 & -0.16 & 1.11 \\
Qwen-VL 2.5 & 8.0 & 1.26 & 1.58 & -0.72 & 1.40 & 0.93 & 1.13 & -0.45 & 1.04 & 0.87 & 1.08 & -0.12 & 1.08 \\
Intern-VL 3 & 8.0 & 1.23 & 1.49 & -1.03 & 1.08 & 1.01 & 1.24 & -0.82 & 0.93 & 0.76 & 0.95 & -0.42 & 0.85 \\
MiniCPM-2.6 & 8.0 & 2.77 & 12.14 & -1.79 & 12.01 & 4.45 & 12.96 & -3.43 & 12.50 & 8.56 & 14.85 & -8.43 & 12.23 \\
BLIP-2 & 8.0 & - & - & - & - & - & - & - & - & - & - & - & - \\
LLaMA-3.2 & 11.0 & 2.39 & 3.00 & -2.19 & 2.06 & 1.74 & 2.29 & -1.53 & 1.71 & 1.48 & 2.13 & -1.11 & 1.82 \\
Intern-VL 3 & 15.0 & 1.24 & 1.55 & -0.85 & 1.29 & 1.03 & 1.29 & -0.47 & 1.20 & 0.89 & 1.11 & -0.18 & 1.10 \\
Qwen-VL 2.5 & 33.0 & 0.98 & 1.28 & -0.31 & 1.24 & 0.89 & 1.17 & -0.10 & 1.17 & 0.89 & 1.14 & 0.39 & 1.07 \\
Intern-VL 3 & 38.0 & 0.89 & 1.08 & -0.58 & 0.92 & 0.97 & 1.17 & -0.68 & 0.96 & 0.98 & 1.21 & -0.38 & 1.15 \\
Qwen A22 & 235.0 & 1.23 & 1.49 & 0.54 & 1.38 & 1.23 & 1.44 & 0.92 & 1.10 & 1.40 & 1.62 & 1.26 & 1.01 \\

Gemini 2.5 Flash & ? & 1.74 & 2.03 & 1.38 & 1.49 & 2.36 & 2.64 & 2.33 & 1.24 & 2.83 & 3.05 & 2.80 & 1.22 \\
Gemini 2.5 Pro & ? & 1.47 & 1.75 & 0.69 & 1.61 & 1.95 & 2.20 & 1.90 & 1.10 & 2.66 & 2.86 & 2.64 & 1.08 \\
Gemini 3 Flash & ? & 1.19 & 1.46 & 0.46 & 1.38 & 1.00 & 1.19 & 0.75 & 0.92 & 1.21 & 1.41 & 1.08 & 0.90 \\
Gemini 3 Pro & ? & 1.00 & 1.31 & -0.36 & 1.26 & 0.74 & 0.99 & 0.18 & 0.97 & 0.90 & 1.12 & 0.70 & 0.87 \\

\bottomrule
\end{tabular}%
}
\caption{Regression metrics for Task-7: Temperature Estimation at varying depth on ThermEval dataset. ↓ indicates lower is better. *Bias closer to 0 is better.}
\end{table*}

\subsubsection{\textbf{Task 6: Temperature Estimation}}

\leavevmode

\begin{figure}[t]
    \centering
    \includegraphics[width=1\linewidth]{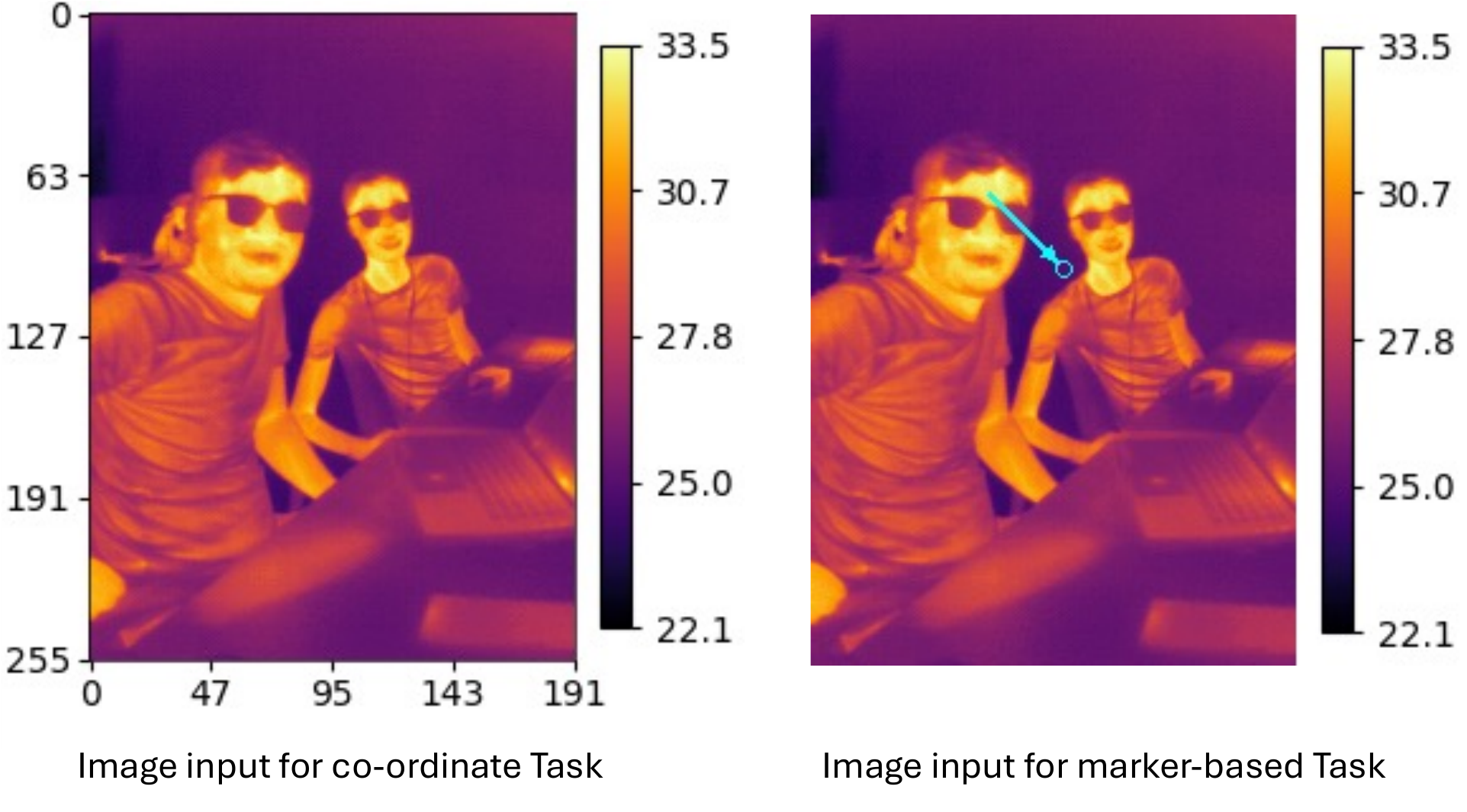}
    \caption{Illustration of how a thermal image gets augmented for co-ordinate and arrow/marker based subtask}
    \label{fig:Task6-illustration}
\end{figure}

\noindent \textbf{Task:} This task analyzes the model's ability to estimate the temperature of given pixels or regions using the colorbar in the image. It is sub-divided into 3 subtasks- (a) Given the coordinates, the model is prompted to estimate the temperature of the given pixel, (b) The model is prompted to estimate the temperature of the pixel marked by a red arrow and (c) The model is required to estimate the temperature of semantic regions like the head, chest or the nose.

\noindent \textbf{Zero-shot prompt subtask 1:}  Given the thermal image, what is the temperature at the coordinates (\{x\},\{y\})? The temperature scale is in degrees Celsius. Strictly return a single numerical value rounded to one decimal place?

\noindent \textbf{Zero-shot prompt subtask 2:}  Given the thermal image, what is the temperature of the pixel at the center of the cyan ring indicated by the cyan arrow? The temperature scale is in degrees Celsius. Strictly return a single numerical value rounded to one decimal place.

\noindent \textbf{Zero-shot prompt subtask 3 a):} Given the thermal image, what is the temperature estimate of the \{body\_part\} according to the image? The temperature scale is in degrees Celsius. Strictly return a single numerical value rounded to one decimal place.

\noindent \textbf{Zero-shot prompt subtask 3 b):}  Given the thermal image, what is the temperature estimate of the \{body\_part\} of the \{right/left\} person according to the image? The temperature scale is in degrees Celsius. Strictly return a single numerical value rounded to one decimal place.

\noindent \textbf{Implementation Details:} All three subtasks are regression tasks using the ThermEval-D dataset, with temperature ground truths obtained via mean of polygon segmentation of temperatures. For the first two subtasks, the coordinates were generated randomly, constrained to the central region of the images to avoid excessive background representation or overlapping with the temperature scale. In the second subtask, the angle of the arrow marking the pixel was also randomized. An illustration of image has been shown in figure~\ref{fig:Task6-illustration}

\subsubsection{\textbf{Task 7: Temperature Estimation at varying distance}}

\leavevmode

\noindent \textbf{Task:} This task analyzes the model's ability to estimate the temperature of given pixels or regions using the colorbar in the image. unlike previous task it is devided by varying distances of 1m , 4m and 6. The model is required to estimate the temperature of semantic regions like the head, chest or the nose.

\noindent \textbf{Zero-shot prompt:} Given the thermal image, what is the temperature estimate of the \{body\_part\} according to the image? The temperature scale is in degrees Celsius. Strictly return a single numerical value rounded to one decimal place.

\noindent \textbf{Implementation Details:} Same as that of Task-6.

\subsection{\textbf{Supervised Finetuning Experiment}}
\label{SFT}

\subsubsection{\textbf{Experimental Setup}}

We fine tuned Qwen2.5 VL 8B Instruct for 5 epochs using LoRA with rank $(r = 16)$ and scaling factor $(\alpha = 16)$, applied to the query, key, value, and output projection layers with dropout (0.1). We used Paged AdamW 32 bit with a fixed learning rate $(5 \times 10^{-6})$, no warmup, batch size 4 per device. The dataset was split into three stratified folds to balance tasks and subtasks, ensuring each VQA sample was seen exactly once and enabling full dataset evaluation without repetition.

\subsubsection{Findings}

Our results show that finetuning Qwen-VL 2.5 enables the model to outperform the much larger Qwen A22 235B and all other evaluated open- and closed-source VLMs. The finetuned model matches or exceeds human performance on most tasks, demonstrating that targeted supervision is highly effective for thermal reasoning. These results indicate that ThermEval provides meaningful, domain-grounding supervision and establishes a reliable benchmark for advancing thermal understanding in VLMs.
\begin{itemize}[noitemsep,topsep=0pt,leftmargin=1.75em]
\item \textbf{Finetuning improves performance but does not solve thermal reasoning.} SFT reduces MAE and improves accuracy, indicating that VLMs have the latent capacity to handle thermal data. However, the remaining errors are still substantial: absolute temperature estimates deviate by 1–2 °C in T6/T7, and performance on semantic comparison tasks (T5) remains below human-level. For applications such as fever screening or industrial hotspot detection, such error margins are not acceptable and highlight the need for deeper physical grounding. For instance, a standard deviation of 1-2 °C would be unsuitable for any model intended for non-contact fever detection.
\item \textbf{SFT closes the gap because current VLMs lack domain grounding, not capacity.} VLMs do not inherently understand temperature as a physical quantity or thermal appearance as a modality, even though they can acquire these concepts with minimal supervision. The fact that small-scale finetuning resolves failures on basic tasks suggests that the primary limitation lies in incomplete training signals rather than in model architecture or scale.
\item \textbf{ThermEval isolates primitive abilities that RGB-centric pretraining does not teach.} Pretrained VLMs, which are predominantly exposed to RGB photographs, diagrams, and charts, tend to learn mappings from appearance to semantic categories. Thermal understanding, however, requires mapping appearance to a physical quantity such as temperature. Because current models are not trained on this type of signal, they do not naturally acquire it, and SFT can only partially bridge this gap.
\item F\textbf{uture VLM pretraining should include physical sensor modalities.} Most existing VLMs are trained primarily on RGB imagery, and although the training data of closed-source models are not public, available evidence and model behavior suggest limited exposure to thermal infrared data. This likely contributes to why current models interpret thermal images as RGB-like visuals rather than as physical measurements. Recent efforts, such as the Gemini team’s inclusion of modalities like X-rays and CT scans, and similar advances in remote-sensing VLMs, demonstrate that expanding pretraining beyond RGB is both feasible and beneficial. These developments indicate that incorporating additional physical sensing modalities is an important direction for future VLM development.
\end{itemize}

\subsection{\textbf{Experimenting with Detailed Prompts}} \label{prompt_ablation}

In these experiments, we added contextual information about the thermal images to guide the VLM toward the relevant aspects of the scene. We conducted a systematic study across Intern-VL (14B), MiniCPM (8B), Qwen-VL-2.5 (7B), and BLIP2 (9B), comparing the original zero-shot prompts with context-augmented versions.

\subsubsection{\textbf{Ablations}}

We describe below the changes we introduced to evaluate the effect of prompt ablations.\\

\noindent \textbf{Task T1 \& T2 : Modality Identification}

\noindent \textbf{Detailed prompt:} ``RGB images come from visible light and depict natural color and texture. Each pixel represents the intensity of red, green and blue channels that together form the visual appearance of objects under illumination. Thermal images measure emitted infrared radiation and encode temperature dependent signals. Each pixel represents a temperature value or a value proportional to heat emission, and any colors seen in the image come from a colormap applied to these underlying temperature readings Based on above context is the given image an RGB or Thermal Image?"\\

\noindent \textbf{Task T3 : Human Counting}

\noindent \textbf{Detailed prompt::} ``You are given a thermal images. Thermal images measure emitted infrared radiation and encode temperature dependent signals. Each pixel represents a temperature value or a value proportional to heat emission, and any colors seen in the image come from a colormap applied to these underlying temperature readings.Count the number of humans in the image."\\

\noindent \textbf{Task T4 : Colorbar Identification}

\noindent \textbf{Detailed prompt:} ``Thermal images encode temperature dependent signals. Each pixel represents a temperature value or a value proportional to heat emission, and any colors seen in the image come from a colormap applied to these underlying temperature readings. They may optionally include a color bar or temperature scale that visually maps colormap colors to corresponding temperature values. Such scales are typically placed along an edge of the image and indicate numeric temperature readings associated with the color gradient. Based on this context, does the given thermal image contain a color bar or temperature scale? Answer only with Yes or No."\\

\noindent \textbf{Task T4 : Colorbar Position Detection}

\noindent \textbf{Detailed prompt:} ``Thermal images encode temperature dependent signals. Each pixel represents a temperature value or a value proportional to heat emission, and any colors seen in the image come from a colormap applied to these underlying temperature readings. They include a color bar or temperature scale that visually maps colormap colors to corresponding temperature values. Such scales are typically placed along an edge of the image and indicate numeric temperature readings associated with the color gradient. Based on this context, determine the location of the color bar in the given thermal image. Possible locations are top, left, bottom, or right."\\

\noindent \textbf{Task T4 : Colorbar Min/Max Extraction}

\noindent \textbf{Detailed prompt:} ``Thermal images encode temperature dependent signals. Each pixel represents a temperature value or a value proportional to heat emission, and any colors seen in the image come from a colormap applied to these underlying temperature readings. They include a color bar or temperature scale that visually maps colormap colors to corresponding temperature values. Such scales are typically placed along an edge of the image and indicate the temperature range represented by the colormap. Using this definition, determine the maximum temperature value shown on the color bar in the given thermal image, expressed in degree Celsius." 

\noindent \textbf{Detailed prompt:} ``Thermal images encode temperature dependent signals. Each pixel represents a temperature value or a value proportional to heat emission, and any colors seen in the image come from a colormap applied to these underlying temperature readings. They include a color bar or temperature scale that visually maps colormap colors to corresponding temperature values. Such scales are typically placed along an edge of the image and indicate the temperature range represented by the colormap. Using this definition, determine the minimum temperature value shown on the color bar in the given thermal image, expressed in degree Celsius."\\

\noindent \textbf{Task T5 : Thermal Reasoning}

\noindent \textbf{Detailed prompt:} ``Thermal images encode temperature dependent signals. Each pixel represents a temperature value or a value proportional to heat emission, and any colors seen in the image come from a colormap applied to these underlying temperature readings. They include a color bar or temperature scale that visually maps colormap colors to corresponding temperature values. Such scales are typically placed along an edge of the image and indicate the temperature range represented by the colormap, enabling comparison of temperatures across different regions. Using this definition, determine which person’s bodypart is hotter in the given thermal image. Respond only with left or right."

\noindent \textbf{Detailed prompt:} ``Thermal images encode temperature dependent signals. Each pixel represents a temperature value or a value proportional to heat emission, and any colors seen in the image come from a colormap applied to these underlying temperature readings. They include a color bar or temperature scale that visually maps colormap colors to corresponding temperature values. Such scales are typically placed along an edge of the image and indicate the temperature range represented by the colormap, enabling comparison of temperatures across different regions. Using this definition, rank the chest, forehead and nose in the given thermal image from highest to lowest temperature. List them from hottest to coolest."\\

\noindent \textbf{Task T6 \& T7  : Temperature Estimation}

\noindent \textbf{Detailed prompt:} ``Thermal images encode temperature dependent infrared signals. Each pixel represents a temperature value or a value proportional to emitted heat, and any visible colors come from a colormap applied to these values. A visible color bar or temperature scale maps colormap colors to numeric temperature readings in degrees Celsius. Given image pixel coordinates (\{x\},\{y\}) with origin at the top-left, x increasing to the right and y increasing downward, report the temperature at the specified coordinates as a single numeric value in degrees Celsius rounded to one decimal place"

\noindent \textbf{Detailed prompt:} ``Thermal images encode temperature dependent infrared signals. Each pixel represents a temperature value or a value proportional to emitted heat, and any visible colors result from a colormap mapped to those values. A visible color bar or temperature scale maps colormap colors to numeric temperature readings in degrees Celsius. The arrow marks a single point in the image. Using the color bar mapping report the temperature at the arrowed point as a single numeric value in degrees Celsius rounded to one decimal place"

\noindent \textbf{Detailed prompt:} ``Thermal images encode temperature dependent infrared signals. Each pixel represents a temperature value or a value proportional to emitted heat, and any visible colors result from a colormap applied to these values. A color bar or temperature scale maps the colormap to numeric temperatures in degrees Celsius, enabling temperature estimation for specific regions. Using this definition, estimate the temperature of the \{body\_part\} in the given thermal image and return a single numeric value in degrees Celsius rounded to one decimal place"

\noindent \textbf{Detailed prompt:} ``Thermal images encode temperature dependent infrared signals. Each pixel represents a temperature value or a value proportional to emitted heat, and any visible colors result from a colormap applied to these values. A color bar or temperature scale maps the colormap to numeric temperatures in degrees Celsius, enabling temperature estimation for specific regions. Using this definition, estimate the temperature of the \{body\_part\} of the \{side\} person in the given thermal image and return a single numeric value in degrees Celsius rounded to one decimal place."

\subsubsection{\textbf{Key Findings}}
Our ablation reveals three clear trends 
\begin{itemize}[leftmargin=1.75em]
    \item Models with reasonable visual grounding (InternVL-14B, MiniCPM, Qwen-VL-2.5) show large gains for simple tasks when contextual modality descriptions are added.
    \begin{itemize}
        \item Qwen-VL-2.5 --- T1 FLIR: 0.71 $\rightarrow$ 0.96 and LLVIP: 0.72 $\rightarrow$ 0.96
        \item Qwen-VL-2.5 --- T2 FLIR: 0.61 $\rightarrow$ 0.98 and LLVIP: 0.80 $\rightarrow$ 0.99
        \item InternVL-14B --- T2 FLIR: 0.86 $\rightarrow$ 0.99 and LLVIP: 0.97 $\rightarrow$ 1.00
    \end{itemize}
    \textbf{Interpretation:} Architecture is not the bottleneck; a short textual description helps models \emph{anchor} the visual signal and correctly name the modality.\\
    \item Across temperature-comparison and temperature-estimation tasks, gains are small, inconsistent, or negative.
    \begin{itemize}
        \item Qwen T6 Arrow: 4.75 $\rightarrow$ 4.46 (small improvement)
        \item MiniCPM T6 Coordinate: 4.00 $\rightarrow$ 4.85 (worse)
        \item InternVL T7 (2ft): 1.01 $\rightarrow$ 2.04 (worse)
    \end{itemize}
    \textbf{Interpretation:} Thermal-physics descriptions in prompts cannot replace the sensor-level priors needed to map pixel or colormap values to temperature. The underlying challenge is the lack of thermal-domain grounding in model pretraining.\\
    \item BLIP-2 often degrades when given extra context:
    \begin{itemize}
        \item T1 FLIR: 0.46 $\rightarrow$ 0.34
        \item T2 FLIR: 0.77 $\rightarrow$ 0.67
        \item T6/T7: unchanged and poor
    \end{itemize}
    \textbf{Interpretation:} When a model lacks basic thermal–visual alignment, prompt engineering cannot compensate for that gap. This aligns with the reviewer intuition that prompting alone is insufficient for reliable thermal reasoning.
\end{itemize}

\begin{figure}[h]
    \centering
    \includegraphics[width=1.0\linewidth]{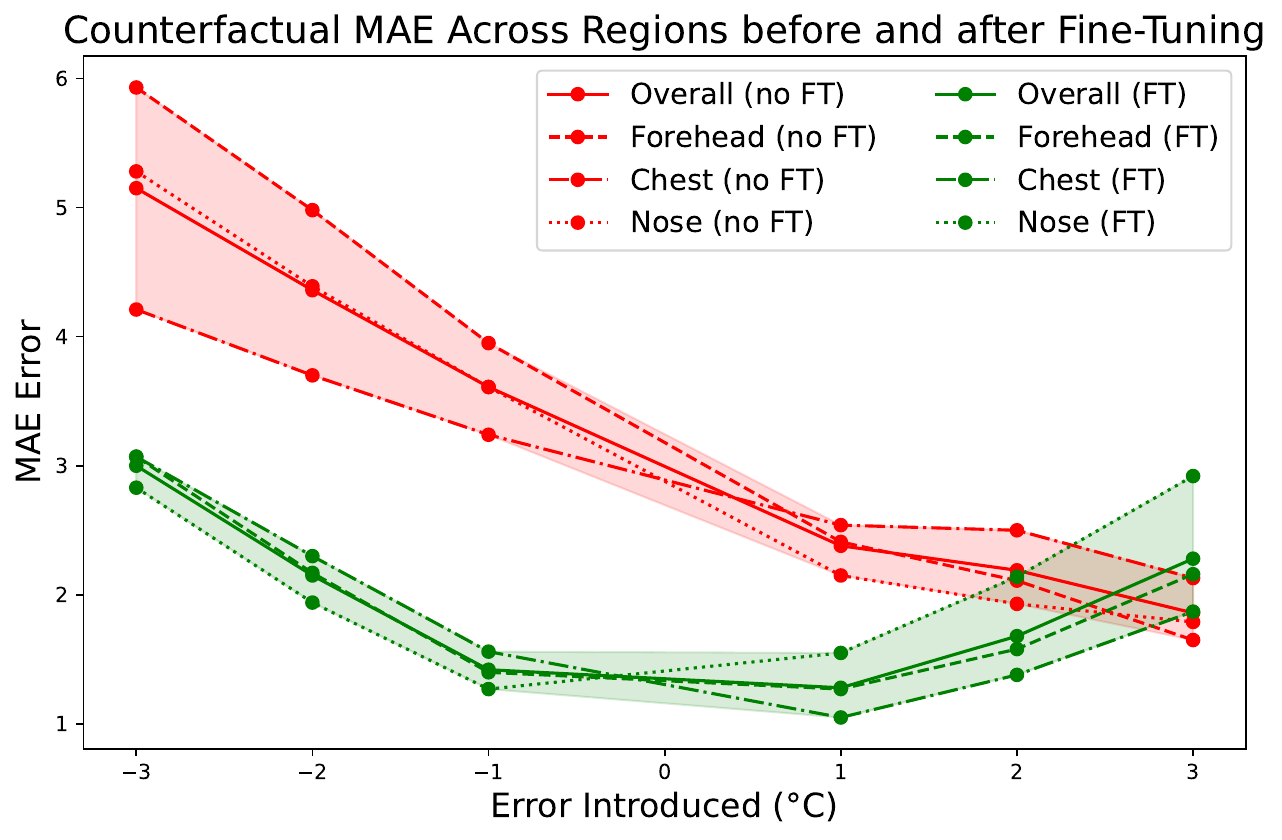}
    \caption{MAE of counterfactual temperature estimation vs. perturbation across regions. Red and green curves represent the model before and after fine-tuning, respectively. "Overall" denotes the aggregate across forehead, chest, and nose.}
    \label{fig:counterfactual}
\end{figure}

\begin{table}[h]
\centering
\caption{Comparison of MAE Results: Without Fine-Tuning vs. After Fine-Tuning}
\resizebox{\linewidth}{!}{
\begin{tabular}{c|cc|cc|cc|cc}
\toprule
\multirow{2}{*}{\textbf{Error}} & \multicolumn{2}{c}{\textbf{OVERALL}} & \multicolumn{2}{c}{\textbf{FOREHEAD}} & \multicolumn{2}{c}{\textbf{CHEST}} & \multicolumn{2}{c}{\textbf{NOSE}} \\ \cline{2-9} 
 & \multicolumn{1}{c}{\textbf{No FT}} & \textbf{After FT} & \multicolumn{1}{c}{\textbf{No FT}} & \textbf{After FT} & \multicolumn{1}{c}{\textbf{No FT}} & \textbf{After FT} & \multicolumn{1}{c}{\textbf{No FT}} & \textbf{After FT} \\ 
 \bottomrule
1  & \multicolumn{1}{c}{2.38} & 1.28 & \multicolumn{1}{c}{2.41} & 1.27 & \multicolumn{1}{c}{2.54} & 1.05 & \multicolumn{1}{c}{2.15} & 1.55 \\ 
2  & \multicolumn{1}{c}{2.19} & 1.68 & \multicolumn{1}{c}{2.11} & 1.58 & \multicolumn{1}{c}{2.50} & 1.38 & \multicolumn{1}{c}{1.93} & 2.14 \\ 
3  & \multicolumn{1}{c}{1.86} & 2.28 & \multicolumn{1}{c}{1.65} & 2.16 & \multicolumn{1}{c}{2.13} & 1.87 & \multicolumn{1}{c}{1.79} & 2.92 \\ 
-1 & \multicolumn{1}{c}{3.61} & 1.42 & \multicolumn{1}{c}{3.95} & 1.40 & \multicolumn{1}{c}{3.24} & 1.56 & \multicolumn{1}{c}{3.61} & 1.27 \\ 
-2 & \multicolumn{1}{c}{4.36} & 2.15 & \multicolumn{1}{c}{4.98} & 2.17 & \multicolumn{1}{c}{3.70} & 2.30 & \multicolumn{1}{c}{4.39} & 1.94 \\ 
-3 & \multicolumn{1}{c}{5.15} & 3.00 & \multicolumn{1}{c}{5.93} & 3.07 & \multicolumn{1}{c}{4.21} & 3.07 & \multicolumn{1}{c}{5.28} & 2.83 \\ 
\bottomrule
\end{tabular}
}
\end{table}

\subsection{Counterfactual Evaluation via Temperature Perturbations}
\label{sec:counterfactuals}

To probe whether VLMs truly ground temperature estimation in thermal cues rather than relying on language or dataset priors, we perform a counterfactual evaluation based on controlled temperature perturbations. For each thermal image, we synthetically alter the pixel values within predefined regions of interest (forehead, chest, and nose) by adding or subtracting a fixed offset of $\pm 1$, $\pm 2$, or $\pm 3$ degrees, while keeping the visual context and temperature scale unchanged. The task is to estimate the absolute temperature of each region.

\noindent We evaluate Qwen2.5-VL-8B under controlled counterfactual temperature perturbations before and after fine-tuning. Without fine-tuning, prediction errors decrease for positive perturbations ($+1$, $+2$, $+3$) and increase sharply for negative ones($-1, -2, -3$), revealing a systematic bias toward higher temperature estimates and limited sensitivity to the direction or magnitude of the perturbation. Predictions collapse to a small set of preferred values, indicating reliance on language or scale priors rather than pixel-level thermal evidence. Fine-tuning restores a more causally consistent response, with errors increasing with perturbation magnitude in both directions. However, the response remains weak, with limited variation across perturbation levels, suggesting residual bias toward dataset-central temperatures. These results show that fine-tuning improves counterfactual sensitivity but does not fully resolve the challenge of precise temperature reasoning from thermal imagery.